\documentclass{article}
\usepackage[preprint]{neurips_2025}
\usepackage{pifont}% http://ctan.org/pkg/pifont
\newcommand{\xmark}{\ding{55}}%
\usepackage[utf8]{inputenc} % allow utf-8 input
\usepackage[T1]{fontenc}    % use 8-bit T1 fonts
\usepackage{hyperref}       % hyperlinks
\usepackage{url}            % simple URL typesetting
\usepackage{booktabs}       % professional-quality tables
\usepackage{amsfonts}       % blackboard math symbols
\usepackage{nicefrac}       % compact symbols for 1/2, etc.
\usepackage{microtype}      % microtypography
\usepackage{xcolor}         % colors
\usepackage{amsmath}
\usepackage{wrapfig}
\usepackage{graphicx}
\usepackage{tikz}
\usetikzlibrary{calc}

\usepackage{placeins}
\usepackage{multirow}   
\usepackage{array}

\usepackage{float}
\usepackage{listings}
\usepackage{courier}
\usepackage{subcaption}
\usepackage{adjustbox}
\usepackage{algorithm}
\usepackage{algpseudocode}

\usepackage{todonotes}
\usepackage{multicol}
\definecolor{forestgreen}{RGB}{34,139,34}
\definecolor{codegreen}{rgb}{0,0.6,0}
\definecolor{codegray}{rgb}{0.5,0.5,0.5}
\definecolor{codepurple}{rgb}{0.58,0,0.82}
\definecolor{backcolour}{rgb}{0.95,0.95,0.92}

\lstdefinestyle{mystyle}{
    backgroundcolor=\color{backcolour},   
    commentstyle=\color{codegreen},
    keywordstyle=\color{magenta},
    numberstyle=\tiny\color{codegray},
    stringstyle=\color{codepurple},
    basicstyle=\small\ttfamily,
    breakatwhitespace=false,         
    breaklines=true,                 
    captionpos=b,                    
    keepspaces=true,                 
    numbers=left,                    
    numbersep=5pt,                  
    showspaces=false,                
    showstringspaces=false,
    showtabs=false,                  
    tabsize=2
}
\usepackage{todonotes}
\usepackage{multicol}

\definecolor{codegreen}{rgb}{0,0.6,0}
\definecolor{codegray}{rgb}{0.5,0.5,0.5}
\definecolor{codepurple}{rgb}{0.58,0,0.82}
\definecolor{backcolour}{rgb}{0.95,0.95,0.92}

\lstdefinestyle{mystyle}{
    backgroundcolor=\color{backcolour},   
    commentstyle=\color{codegreen},
    keywordstyle=\color{magenta},
    numberstyle=\tiny\color{codegray},
    stringstyle=\color{codepurple},
    basicstyle=\small\ttfamily,
    breakatwhitespace=false,         
    breaklines=true,                 
    captionpos=b,                    
    keepspaces=true,                 
    numbers=left,                    
    numbersep=5pt,                  
    showspaces=false,                
    showstringspaces=false,
    showtabs=false,                  
    tabsize=2
}
\let\ACMmaketitle=\maketitle
\renewcommand{\maketitle}{\begingroup\let\footnote=\thanks \ACMmaketitle\endgroup}
\title{\Large{Myosotis: structured computation for attention like layer}}
\author{
Evgenii Egorov
\thanks{\small{Work done during an internship in Qualcomm AI Research.~Alt. email: \tiny{\texttt{email.evgenii.egorov@gmail.com}}}}\\
University of Amsterdam \\
\small{\texttt{egorov.evgenyy@ya.ru}}
\and
\textbf{Hanno Ackermann}
\quad\textbf{Markus Nagel}
\quad\textbf{Hong Cai} \\
Qualcomm AI Research\thanks{\small{Qualcomm AI Research is an initiative of Qualcomm Technologies, Inc.}}\\
\small{\texttt{hackerma, markusn, hongcai @qti.qualcomm.com}}\\ 
}
\begin{document}
\maketitle

\begin{abstract}
Attention layers apply a sequence-to-sequence mapping whose parameters depend on the pairwise interactions of the input elements.
However, without any structural assumptions, memory and compute scale quadratically with the sequence length. The two main ways to mitigate this are to introduce sparsity by ignoring a sufficient amount of pairwise interactions or to introduce recurrent dependence along them, as SSM does. Although both approaches are reasonable, they both have disadvantages. 
We propose a novel algorithm that combines the advantages of both concepts. Our idea is based on the efficient inversion of tree-structured matrices. 
\end{abstract}

\section{Introduction}
Modeling interactions in high-dimensional objects efficiently has been a long-standing challenge in machine learning, particularly when short-range dependencies are insufficient and long-range interactions must be captured. The Transformer architecture \cite{vaswani2017attention} was a breakthrough in addressing long-range dependencies, but it suffers from quadratic computational complexity with respect to input length. To mitigate this, numerous efficient Transformer variants have been proposed, exploiting sparsity or hierarchical structure. Examples include BigBird \cite{zaheer2020bigbird}, Longformer \cite{beltagy2020longformer}, Performer \cite{choromanski2021rethinking}, and Swin Transformer \cite{liu2021Swin}, each designed to handle longer sequences or high-dimensional inputs.

An alternative approach is offered by State-Space Models (SSMs), which recast sequence modeling as linear dynamical systems, allowing exact or approximate recurrent computation over long sequences \cite{gu2020hippo, gu2021combining, gu2022efficiently, gu2023hippo}. SSMs have been extended to multidimensional signals, audio and video tasks, demonstrating strong performance while maintaining linear complexity \cite{nguyen2022s4nd, goel2022sashimi}. Variants like S4 \cite{gu2021s4} and S5\cite{smith2023simplified} further improve parameterization and initialization, leading to robust training and effective long-range dependency modeling.

Building on advances in graphical models and deep learning architectures, we introduce a new structural state-space layer, Myo, illustrated in Fig. \ref{fig:myostructure}. Myo retains the linear complexity of existing SSM layers but explicitly leverages structural assumptions about token interactions, connecting sparsity patterns with recurrence constraints via matrix inversion. This framework allows SSM layers to emerge as a special case and enables direct inheritance of initialization schemes from the SSM literature \cite{gu2022s4d, gu2023hippo}. The layer is straightforward to implement (see Sec. \ref{sec:app_pseudocode}) and can incorporate domain knowledge in the form of graphs.

\section{Background}
In this section, we review two related sequence-to-sequence layers: the state space model layer (SSM) and self-attention. As our goal is computational complexity, we focus on the core components of the layer and ignore mapping from input sequence to the parameters of the layer. We identify a gap between two approaches that motivates our solution.

We denote tuples of length $L$ by $s_{1:L}$ or $s$, when the length is clear. If all elements of the sequence belong to the same space $V$, then we write $s\in V^{L}$, otherwise specify per element $s_k\in V_{k}$. The first index of sequence is $1$. We denote the input sequence of the layer as a tuple $u\in\mathbb{R}^{L\times M}$ and the output sequence as a tuple $x\in\mathbb{R}^{L\times N}$. We index elements of a tuple with lower index, for example, $u_t\in\mathbb{R}^{M}, x_k\in\mathbb{R}^{N}$ are the $t$-th and the $k$-th elements of the sequences $x$ and $u$. 
Both self-attention and state space layers are parametrized of the sequence-to-sequence mapping $K: \mathbb{R}^{L\times M}\to \mathbb{R}^{L\times N}$, which itself can be input dependent. We refer to its matrix as the kernel matrix.

\subsection{State Space Model (SSM) Layer}
Given an input sequence $u_{1:L}$ and parameter sequences $A_{1:L}$ and $B_{1:L}$, with $A_k\in\mathbb{R}^{N\times N}$, $B_k\in\mathbb{R}^{N \times M}$.
A state space layer maps an input sequence to the output $x_{1:L}$ by the following recurrence:
\begin{equation}
    \begin{aligned}
    \label{eq:rec}
    %& x_{1} = B_1u_1, & k\in\{2,\dots, L\}: x_k = A_{k-1}x_{k-1} + B_{k}u_k.
    x_{1} &= B_1u_1, & x_k &= A_{k-1}x_{k-1} + B_{k}u_k \quad \mbox{for} \; k\in\{2,\dots, L\}.
    \end{aligned}
\end{equation}
As matrix multiplication is an associative operation $((AB)(CD) = ((A(B(C(D))))$, the above recurrence can be computed in parallel by reusing intermediate computations using \textit{associative scan} operations. As differentiation is a linear operation, the same is applicable for differentiation through the associative scan on a backward pass. The time complexity on $T$ processors is $O((L/T + \log_2 T)$ and the space complexity is linear over the sequence length $L$.
\subsection{Attention Layer}
We are given the input sequence $u_{1:L}$ and parameter sequences $q_{1:L}, k_{1:L}$, $q_k\in\mathbb{R}^{N}$, $k_k\in\mathbb{R}^{N}$ and a nonlinearity $\sigma$. We call the composition of the non-linearity and inner product the kernel $k(x,y)= \sigma(\langle x, y\rangle)$ of the attention. The attention layer maps the input sequence $u_{1:L}$ to the output sequence $x_{1:L}$ as follows:
\begin{equation}
    \begin{aligned}
    & k \in \{1,\dots, L\},
    & x_k = \sum_{n=1}^{L} u_n \cdot \dfrac{\sigma(\langle q_{k}, k_{n}\rangle)}{\sum_{n'\in\{1,\dots,L\}}\sigma(\langle q_{k}, k_{n'}\rangle)}. 
    \end{aligned}
\end{equation}
Hence, the complexity over both time and space of the self-attention layer is quadratic with respect to the length of the sequence $L$. To reduce it, additional assumptions about a function $k(x, y)$ should be made. Two common approaches are the separability and sparsity assumptions.
\paragraph{Separability assumption} 
We consider the following parameterization of a fixed or learnable dictionary of functions $\{\phi_p\}_{p=1}^{P},\phi_p:\mathbb{R}^{N}\to\mathbb{R}^{N_p}$. 
Using it, we define a separable kernel:
\begin{equation}
    \begin{aligned}
        & k(x, y) = \sum_{p=1}^{P}\langle \phi_{p}(x), \phi_{p}(y) \rangle, x,y \in \mathbb{R}^N.
    \end{aligned}
\end{equation}
As $\sum_{p=1}^{P}\langle\phi_{p}(x), \phi_{p}(y)\rangle = \langle (\phi_{1}(x), \dots, \phi_{P}(x)), (\phi_{1}(y), \dots, \phi_{P}(y)) \rangle$, we consider the feature stacking map $\psi: \mathbb{R}^{N} \to \mathbb{R}^{\sum_{n=1}^P N_p}, \psi(x) = (\phi_{1}(x), \dots, \phi_{P}(x)))$ and hence the "linear" self-attention map:
\begin{equation}
    %\begin{aligned}
    %& k \in \{1,\dots, L\}, Q_k = \psi(q_k) \in \mathbb{R}^{\sum_{n=1}^P N_p}, K_{k} = \psi(k_{k}) \in \mathbb{R}^{D}, D = \sum_{n\in\{1,\dots,P\}}N_p, \\
    Q_k = \psi(q_k) \in \mathbb{R}^D, \;\; K_{k} = \psi(k_{k}) \in \mathbb{R}^{D}, \;\; D = \sum_{n\in\{1,\dots,P\}}N_p \;\; \mbox{for} \;\; k \in \{1,\dots, L\} 
\end{equation}
%\hanno{How does $x_k$ come into the story?}
\begin{equation}
    \begin{aligned}
    x_k = \sum_{n=1}^{L} u_n \cdot \dfrac{\langle Q_{k}, K_{n}\rangle}{\sum_{n'=1}^{L}\langle Q_{k}, K_{n'} \rangle} = (\langle Q_{k},\sum_{n'\in\{1,\dots,L\}} K_{n'} \rangle)^{-1}\sum_{p=1}^{D} \left(\sum_{n=1}^{L} u_n \cdot (K_n)_p \right) (Q_k)_p.
    \end{aligned} 
\end{equation}
As a result, for fixed $P\ll L$, we obtain a desirable linear complexity solution over the length of the sequence. If we keep only the lower triangular part ($n \leq k)$, the relation to the state-space model can be made more explicit. Consider $x_k$ before scaling and summating over $p$ and denote it by $x_k^p$:
\begin{equation}
    \begin{aligned}
    & x^p_{k} = q^p_k\sum_{n \leq k} u_n \cdot k^p_n = \frac{q_k^P}{q_{k-1}^p}x^p_{k-1} + (k^p_k q_k^p) u_k,
    \end{aligned}
\end{equation}
which reassembles the recurrence.

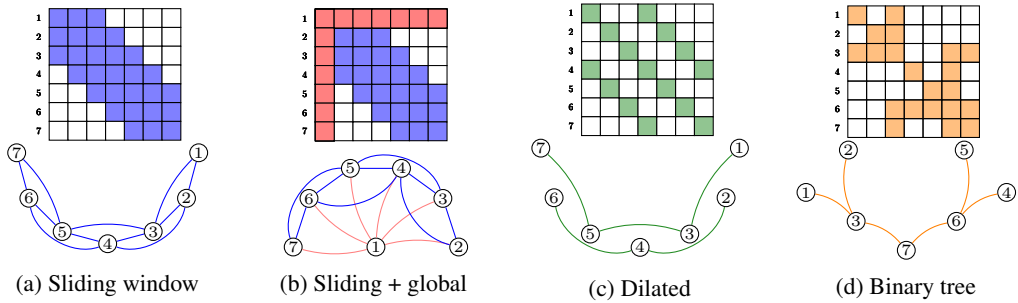
\begin{figure}[H]
    \centering
    \scriptsize % Smaller font for TikZ and captions
\def\n{7} % sequence length
\def\w{2} % window radius
\def\s{3} % dilation stride

\newlength{\figwidth}
\setlength{\figwidth}{0.24\linewidth} % slightly less than 1/4 to fit 4 in a row

\begin{center}

% ---------- Sliding window ----------
\begin{minipage}{\figwidth}
\centering
\begin{tikzpicture}[every node/.style={minimum size=0.5cm, minimum width=0.5cm, align=center}, scale=0.25, transform shape]
    \foreach \i in {1,...,\n} {
        \foreach \j in {1,...,\n} {
            \pgfmathtruncatemacro{\low}{\i-\w}
            \pgfmathtruncatemacro{\high}{\i+\w}
            \node[left, font=\normalsize\Large] at (0.7,-\i-0.5) {$\i$};
            \ifnum\j<\low
                \fill[white] (\j,-\i) rectangle ++(1,-1);
            \else
                \ifnum\j>\high
                    \fill[white] (\j,-\i) rectangle ++(1,-1);
                \else
                    \fill[blue!50] (\j,-\i) rectangle ++(1,-1);
                \fi
            \fi
            \draw[black] (\j,-\i) rectangle ++(1,-1);
        }
    }
\end{tikzpicture}

\begin{tikzpicture}[scale=0.55, every node/.style={circle, draw, minimum size=0mm, inner sep=0.5pt, font=\tiny}]
    \def\n{7}
    \def\r{2.2} % increased radius for balance

    \foreach \i in {1,...,\n} {
        \pgfmathsetmacro{\angle}{360 - 180*(\i-1)/(\n-1)}
        \node (v\i) at ({\r*cos(\angle)}, {\r*sin(\angle)}) {$\i$};
    }
    
    \foreach \i in {2,...,7} {
        \pgfmathtruncatemacro{\a}{\i-1}
        \draw[blue] (v\i) -- (v\a);
    }

    \foreach \i in {2,...,7}{
        \ifnum \i>2
            \pgfmathtruncatemacro{\b}{\i-2}
            \ifodd\i
                \draw[bend left=15, blue, looseness=0.9] (v\i) to (v\b);
            \else
                \draw[bend right=50, blue, looseness=0.9] (v\i) to (v\b);
            \fi
        \fi
    }
\end{tikzpicture}
\subcaption{Sliding window}
\end{minipage}\hfill
%
% ---------- Sliding + global ----------
\begin{minipage}{\figwidth}
\centering
\begin{tikzpicture}[every node/.style={minimum size=0.5cm, minimum width=0.5cm, align=center}, scale=0.25, transform shape]
    \foreach \i in {1,...,\n} {
        \foreach \j in {1,...,\n} {
            \pgfmathtruncatemacro{\low}{\i-\w}
            \pgfmathtruncatemacro{\high}{\i+\w}
            \ifnum\j<\low
                \fill[white] (\j,-\i) rectangle ++(1,-1);
            \else
                \ifnum\j>\high
                    \fill[white] (\j,-\i) rectangle ++(1,-1);
                \else
                    \fill[blue!50] (\j,-\i) rectangle ++(1,-1);
                \fi
            \fi
            \draw[black] (\j,-\i) rectangle ++(1,-1);
        }

        \foreach \i in {1,...,\n} {
            \node[left, font=\normalsize\Large] at (0.7,-\i-0.5) {$\i$};
            \fill[red!50] (1,-\i) rectangle ++(1,-1);
            \fill[red!50] (\i,-1) rectangle ++(1,-1);
            \draw[black] (1,-\i) rectangle ++(1,-1);
            \draw[black] (\i, -1) rectangle ++(1,-1);
        }
    }
\end{tikzpicture}

\begin{tikzpicture}[scale=0.55, every node/.style={circle, draw, minimum size=0mm, inner sep=0.5pt, font=\tiny}]
    \def\n{7}
    \def\r{2.0}

    \foreach \i in {1,...,\n} {
        \pgfmathsetmacro{\angle}{180*(\i-2)/(\n-2)}
        \ifnum\i=1
            \node (v\i) at ({0}, {0}) {$1$};
        \else
            \node (v\i) at ({\r*cos(\angle)}, {\r*sin(\angle)}) {$\i$};
        \fi
    }
    
    \foreach \i in {2,...,7} {
        \pgfmathtruncatemacro{\a}{\i-1}
        \ifnum\a>1
            \draw[blue] (v\i) -- (v\a);
        \fi
        \draw[bend left=15, red!50, looseness=0.9] (v1) to (v\i);
    }

    \foreach \i in {3,...,7}{
        \pgfmathtruncatemacro{\b}{\i-2}
        \ifodd\i
            \ifnum\b>1
                \draw[bend left=55, blue, looseness=0.9] (v\i) to (v\b);
            \fi
        \else
            \draw[bend right=40, blue, looseness=0.9] (v\i) to (v\b);
        \fi
    }
\end{tikzpicture}

\subcaption{Sliding + global}
\end{minipage}\hfill
%
% ---------- Dilated ----------
\begin{minipage}{\figwidth}
\centering
\begin{tikzpicture}[every node/.style={minimum size=0.5cm, minimum width=0.5cm, align=center}, scale=0.25, transform shape]
 \foreach \i in {1,...,\n} {
    \foreach \j in {1,...,\n} {
        \pgfmathtruncatemacro{\diff}{abs(\i-\j)}
        \pgfmathtruncatemacro{\modval}{mod(\diff,\s)}
        \ifnum\diff=0
            \fill[forestgreen!50] (\j,-\i) rectangle ++(1,-1);
        \else
            \ifnum\diff=3
                \fill[forestgreen!50] (\j,-\i) rectangle ++(1,-1);
            \else
                \fill[white] (\j,-\i) rectangle ++(1,-1);
            \fi
        \fi
        \node[left, font=\normalsize\Large] at (0.7,-\i-0.5) {$\i$};
        \draw[black] (\j,-\i) rectangle ++(1,-1);
    }
 }
\end{tikzpicture}

\begin{tikzpicture}[scale=0.55, every node/.style={circle, draw, minimum size=0mm, inner sep=0.5pt, font=\tiny}]
    \def\n{7}
    \def\r{2.4}

    \foreach \i in {1,...,\n} {
        \pgfmathsetmacro{\angle}{360 - 180*(\i-1)/(\n-1)}
        \node (v\i) at ({\r*cos(\angle)}, {\r*sin(\angle)}) {$\i$};
    }
    
    \foreach \i in {2,...,7}{
        \ifnum \i>2
            \pgfmathtruncatemacro{\b}{\i-2}
            \ifodd\i
                \draw[bend left=18, forestgreen, looseness=0.9] (v\i) to (v\b);
            \else
                \draw[bend right=50, forestgreen, looseness=0.9] (v\i) to (v\b);
            \fi
        \fi
    }
\end{tikzpicture}

\subcaption{Dilated}
\end{minipage}\hfill
%
% ---------- Binary tree ----------
\begin{minipage}{\figwidth}
\centering
\begin{tikzpicture}[every node/.style={minimum size=0.5cm, minimum width=0.5cm, align=center}, scale=0.25, transform shape]
\def\edges{1/3,2/3,4/6,5/6,3/6,3/7,6/7}
\foreach \a/\b in \edges{
    \fill[orange!50] (\a,-\b) rectangle ++(1,-1);
    \fill[orange!50] (\b,-\a) rectangle ++(1,-1);
    \fill[orange!50] (\a,-\a) rectangle ++(1,-1);
}
\fill[orange!50] (7,-7) rectangle ++(1,-1);

\foreach \i in {1,...,\n} {
    \foreach \j in {1,...,\n} {
        \draw[black] (\j,-\i) rectangle ++(1,-1);
        \node[left, font=\normalsize\Large] at (0.7,-\i-0.5) {$\i$};
    }
}
\end{tikzpicture}

\begin{tikzpicture}[scale=0.55, every node/.style={circle, draw, minimum size=0mm, inner sep=0.5pt, font=\tiny}]
    \def\hr{1.4}
    \def\angle{180/6}

    \node (v7) at (0,0) {$7$};
    \node (v6) at ({\hr*cos(\angle)}, {\hr*sin(\angle)}) {$6$};
    \node (v3) at ({\hr*cos(-\angle+180)}, {\hr*sin(-\angle+180)}) {$3$};
    
    \node (v2) at ({\hr*2*cos(-2*\angle+180)}, {\hr*2*sin(-2*\angle+180)}) {$2$};
    \node (v1) at ({\hr*2*cos(-1*\angle+180)}, {\hr*2*sin(-1*\angle+180)}) {$1$};
    \node (v4) at ({\hr*2*cos(1*\angle)}, {\hr*2*sin(1*\angle)}) {$4$};
    \node (v5) at ({\hr*2*cos(2*\angle)}, {\hr*2*sin(2*\angle)}) {$5$};

    \draw[bend left=20, orange, looseness=0.9] (v1) to (v3);
    \draw[bend right=20, orange, looseness=0.9] (v2) to (v3);
    \draw[bend right=20, orange, looseness=0.9] (v4) to (v6);
    \draw[bend left=20, orange, looseness=0.9] (v5) to (v6);
    \draw[bend left=20, orange, looseness=0.9] (v3) to (v7);
    \draw[bend right=20, orange, looseness=0.9] (v6) to (v7);
\end{tikzpicture}
\subcaption{Binary tree}
\end{minipage}

\end{center}
    \caption{Graphs of sparsity patterns (without self-loop edges) and corresponding attention matrices. Labels of nodes correspond to the usual enumeration: the top row is $1$ and the last is $7$.}
    \label{fig:attn_sparse_exmp}
\end{figure}

Note that the separability assumption on the kernel reduces the number of free parameters of the kernel matrix $K$ from $L^2$ to $L\times D$. Although the kernel matrix $K$ is generally dense, this reduces computational costs from quadratic to linear over the sequence length.
\paragraph{Sparsity assumptions}
Another way to reduce complexity is to introduce a structural sparsity pattern in the attention kernel computation. It is convenient to represent a sparsity assumption by a graph. Consider a directed graph $G(V, E)$, where $V = \{1,\dots,L\}$ is a set of vertices, and $E$ is a set of directed edges over $V$.  Vertexes of the graph correspond to the sequence elements, and the directed edges are present if we assume a nonzero attention value for a general input. For each pair $(n,m)\in V\times V$, we have the following matrix element:
\begin{equation}
\begin{aligned}
    & k_{nm} = \sigma(q_n, k_m), & \text{if } (n,m)\in E, & \text{ otherwise } 0. 
\end{aligned}
\end{equation}
If the graph has structured sparsity, it can be used to reduce the computation cost. A simple example is a pattern of edges present in sliding windows $E=\{(n, m):~|n - m| \leq T\}$, for some constant $T$. The kernel matrix $K$ will be a band matrix, with block size $T$ and therefore the computational and memory complexity will be $O(TL)$. We illustrate some sparsity patterns and the corresponding attention matrices in Figure \ref{fig:attn_sparse_exmp}.

\section{The Myosotis (Myo) Layer}
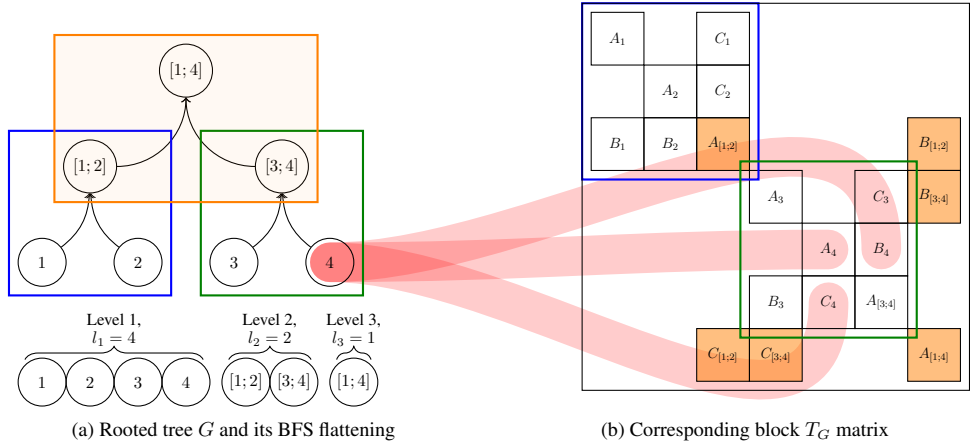
\begin{figure}[h!]
    \centering
    \usetikzlibrary{decorations.text}
\usetikzlibrary{decorations.pathreplacing}
\usetikzlibrary{fit}
\definecolor{darkgreen}{rgb}{0.0, 0.5, 0.0}
\usetikzlibrary{shapes}
\pgfdeclarelayer{bg}
\pgfsetlayers{bg,main}

\begin{tikzpicture}[scale=0.85, transform shape, every node/.style={minimum size=0.5cm, minimum width=0.5cm, align=center}]
\begin{scope}[scale=0.75, transform shape]
\node (1) at (-4,-4) [circle, draw, minimum size=1cm, inner sep=1.5pt] {$1$};
\node (2) at (-2,-4) [circle, draw, minimum size=1cm, inner sep=1.5pt]  {$2$};
\node (3) at (0, -4) [circle, draw, minimum size=1cm, inner sep=1.5pt]  {$3$};
\node (4) at (2, -4) [circle, draw, minimum size=1cm, inner sep=1.5pt]  {$4$};
\node (12) at (-3,-2) [circle, draw] {$[1;2]$};
\node (34) at (1,-2) [circle, draw]  {$[3;4]$};
\node (1234) at (-1, 0) [circle, draw] {$[1;4]$};

\draw[->, bend right=25] (1) to (12);
\draw[->, bend left=25] (2) to (12);
\draw[->, bend right=25] (3) to (34);
\draw[->, bend left=25] (4) to (34);
\draw[->, bend right=45] (12) to (1234);
\draw[->, bend left=45] (34) to (1234);

\node[draw=blue, thick, rectangle, fit=(1) (2) (12)] {};
\node[draw=darkgreen, thick, rectangle, fit=(3) (4) (34)] {};
\node[draw=orange, fill=orange, fill opacity=0.05, thick, rectangle, fit=(12) (34) (1234)] {};

\begin{scope}[shift={(-2,-5)}]
\node (l1) at (-2,-1.5) [draw, circle, minimum size=1cm, inner sep=0] {$1$};
\node (l2) at (-1,-1.5) [draw, circle, minimum size=1cm, inner sep=0]  {$2$};
\node (l3) at (0, -1.5) [draw, circle, minimum size=1cm, inner sep=0]  {$3$};
\node (l4) at (1, -1.5) [draw, circle, minimum size=1cm, inner sep=0]  {$4$};
\node (l12) at (2.25,-1.5) [draw, circle, minimum size=1cm, inner sep=0] {$[1;2]$};
\node (l34) at (3.25,-1.5) [draw, circle, minimum size=1cm, inner sep=0]  {$[3;4]$};
\node (l1234) at (4.5, -1.5) [draw, circle, minimum size=1cm, inner sep=0] {$[1;4]$};

\draw [decorate,decoration={brace,amplitude=5pt,raise=2pt},yshift=-10pt] (l1.north west) -- (l4.north east) node [black,midway,yshift=21pt] {Level 1, \\ $l_1 = 4$};
\draw [decorate,decoration={brace,amplitude=5pt,raise=2pt},yshift=0pt] (l12.north west) -- (l34.north east) node [black,midway,yshift=21pt] {Level 2, \\ $l_2 = 2$};
\draw [decorate,decoration={brace,amplitude=5pt,raise=2pt},yshift=0pt] (l1234.north west) -- (l1234.north east) node [black,midway,yshift=21pt] {Level 3, \\ $l_3 = 1$};
\end{scope}
\end{scope}

\begin{scope}[shift={(6, 0.5)}]
\begin{scope}[scale=0.55, transform shape]
\tikzstyle{tight node} = [draw, minimum size=1.5cm, inner sep=1pt, align=center, font=\large]
  \node[tight node] (m1) at (0, 0) {$A_1$};
  \node[tight node] (m2) at (1.5, -1.5) {$A_2$};
  \node[tight node, fill=orange, fill opacity=0.5, draw opacity=1., text opacity=1., draw=black] (m12) at (3, -3) {$A_{[1;2]}$};
  \node[tight node] (m12_1c) at (0, -3) {$B_{1}$};
  \node[tight node] (m12_1r) at (3, 0) {$C_{1}$};
  \node[tight node] (m12_2c) at (1.5, -3) {$B_{2}$};
  \node[tight node] (m12_2r) at (3, -1.5) {$C_{2}$};
  \node[tight node] (m3) at (4.5, -4.5) {$A_{3}$};
  \node[tight node] (m4) at (6, -6) {$A_{4}$};
  \node[tight node] (m34) at (7.5, -7.5) {$A_{[3;4]}$};
  \node[tight node] (m34_3r) at (4.5, -7.5) {$B_{3}$};
  \node[tight node] (m34_3c) at (7.5, -4.5) {$C_{3}$};
  \node[tight node] (m34_4r) at (7.5, -6) {$B_{4}$};
  \node[tight node] (m34_4c) at (6, -7.5) {$C_{4}$};
  \node[tight node, fill=orange, fill opacity=0.5, draw opacity=1., text opacity=1., draw=black] (m1234_12r) at (9, -3) {$B_{[1;2]}$};
  \node[tight node, fill=orange, fill opacity=0.5, draw opacity=1., text opacity=1., draw=black] (m1234_12c) at (3, -9) {$C_{[1;2]}$};
  \node[tight node, fill=orange, fill opacity=0.5, draw opacity=1., text opacity=1., draw=black] (m1234_34r) at (9, -4.5) {$B_{[3;4]}$};
  \node[tight node, fill=orange, fill opacity=0.5, draw opacity=1., text opacity=1., draw=black] (m1234_34c) at (4.5, -9) {$C_{[3;4]}$};
  \node[tight node, fill=orange, fill opacity=0.5, draw opacity=1., text opacity=1., draw=black] (m1234) at (9, -9) {$A_{[1;4]}$};
\node[draw=blue, thick, rectangle, fit=(m1) (m2) (m12)] {};
\node[draw=darkgreen, thick, rectangle, fit=(m3) (m4) (m34)] {};
\end{scope}
\end{scope}
\tikzset{
  wrap/.style={
    line cap=round,
    #1,
    line width=15pt,
    opacity=0.2,
  },
}
\begin{pgfonlayer}{bg}
    \draw[wrap=red](4.center)   to[out=0,in=180](m4.center);
    \draw[wrap=red](4.center)   to[out=0,in=-90](m34_4c.center);
    \draw[wrap=red](4.center)   to[out=0,in=90](m34_4r.center);
\end{pgfonlayer}
\node at (0, -5.55) {\parbox{0.5\linewidth}{\subcaption{Rooted tree $G$ and its BFS flattening}\label{subfig:a}}};
\node at (8, -5.55) {\parbox{0.5\linewidth}{\subcaption{Corresponding block $T_{G}$ matrix}\label{subfig:b}}};

\node[draw=black, fit=(m1)(m2)(m12)(m12_1c)(m12_1r)(m12_2c)(m12_2r) (m3)(m4)(m34)(m34_3r)(m34_3c)(m34_4r)(m34_4c) (m1234_12r)(m1234_12c)(m1234_34r)(m1234_34c)(m1234)] {};

\end{tikzpicture}
    \vskip -20pt
    \caption{Correspondence between layer representation as a tree structure and block matrix representation. For element $4$ red arrows highlight a tuple $(A_4, C_4, B_4)$ in a graph node and its places in a matrix.}
    \label{fig:myostructure}
\end{figure}
\setlength{\figwidth}{0.45\linewidth}

\subsection{Identifying a gap}
As we can see in Figure \ref{fig:attn_sparse_exmp}, there is a wide variety of sparsity patterns, possibly reflecting domain knowledge. Although this is an efficient way to reduce computational cost, hard-zeroing out elements is a restrictive choice. In contrast, SSM-like layers invoke separability, which keeps the attention matrix dense, yet has an efficient way of computation via recurrence. Intuitively, it is clear that the recurrence in \ref{eq:rec} corresponds to the chain graph. Motivated by this, we make the connection precise and propose a Myosotis\footnote{The forget-me-not flower's scientific name: Myosotis} layer.

The sketch of the approach is the following. We construct a tree graph $G$ from a sequence and consider the corresponding block tree matrix $T_{G}$. The output of the proposed layer is the application of $T_{G}^{-1}$ to the input $u$. As the matrix is tree-structured (block), therefore solving $T_G x = u$ is an efficient recurrence-like computation. Hence, we fuse both ideas discussed above: we use the sparse $T_{G}$, but apply the dense $T_{G}^{-1}$. Next, we discuss the algorithm in detail.

\subsection{Myo Layer: Structure}
For explanation purposes in this section, we will construct the matrix associated with the layer explicitly; however, as we discuss further application of the layer, it is matrix-free. We parameterize the layer with a rooted tree graph $G(V, E)$. We consider a (reversed) breadth-first search traverse (BFS) of the graph $G$, with $D$ levels, where the bottom (leaf) level is the first level and the root node is the last. The number of nodes at each level is denoted by $l_k,k\in\{1,\dots,D\}$, with $L=\sum_{k=1}^{D}l_k$. We label the leaf vertices from $1$ to $l_1$ according to the BFS order. The non-leaf node is labeled by its subtree with segment $[n,m]$ of the most left leaf $n$ and the most right leaf $m$. For example, the root label is labeled $[1;l_1]$ as it covers all leaves.

\paragraph{Layer input and output}
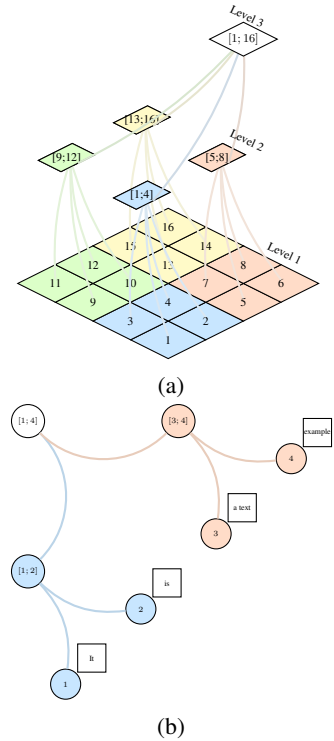
\begin{wrapfigure}{r}{0.35\textwidth} % 'r' for right, adjust width as needed
  \centering
  \begin{subfigure}{\linewidth}
    \centering
    \begin{tikzpicture}[scale=0.5, every node/.style={font=\small, transform shape}]

% Updated isometric projection
\tikzset{iso/.style={x={(1.0cm,0.5cm)}, y={(-1.0cm,0.5cm)}}}

% Colors for each 2x2 block
\definecolor{block0}{RGB}{200,230,255}
\definecolor{block1}{RGB}{255,220,200}
\definecolor{block2}{RGB}{220,255,200}
\definecolor{block3}{RGB}{255,250,200}
\definecolor{minmaxcolor}{RGB}{180,255,180}

\begin{scope}[iso]
  \foreach \x in {0,...,3} {
    \foreach \y in {0,...,3} {
      \pgfmathtruncatemacro{\blockid}{floor(\x/2) + 2*floor(\y/2)}
      \ifcase\blockid
        \def\thiscolor{block0}
      \or
        \def\thiscolor{block1}
      \or
        \def\thiscolor{block2}
      \or
        \def\thiscolor{block3}
      \fi
      \filldraw[fill=\thiscolor, draw=black] (\x,\y) -- ++(1,0) -- ++(0,1) -- ++(-1,0) -- cycle;
      \coordinate (pix-\x-\y) at (\x+0.5,\y+0.5);
      \pgfmathtruncatemacro{\xL}{mod(\x,2)}
      \pgfmathtruncatemacro{\yL}{mod(\y,2)}
      \pgfmathtruncatemacro{\xH}{floor(\x/2)}
      \pgfmathtruncatemacro{\yH}{floor(\y/2)}
      \pgfmathtruncatemacro{\morton}{\xL + 2*\yL + 4*\xH + 8*\yH + 1}
      \node at (pix-\x-\y) {\morton};
    }
  }

  \foreach \xH in {1,0} {
    \foreach \yH in {1,0} {
      \pgfmathtruncatemacro{\gid}{\xH + 2*\yH}
      \coordinate (g\gid-base) at (\xH*2 + 4, \yH*2 + 2);
      \pgfmathtruncatemacro{\minVal}{\gid*4 + 1}
      \pgfmathtruncatemacro{\maxVal}{\minVal + 3}
      \coordinate (top-left) at ($(g\gid-base)+(-0.45,3)$);
      \coordinate (top-right) at ($(g\gid-base)+(0.45,3)$);
      \coordinate (bottom-right) at ($(g\gid-base)+(0.45,2.4)$);
      \coordinate (bottom-left) at ($(g\gid-base)+(-0.45,2.4)$);

      \ifcase\gid
        \def\thiscolor{block0}
      \or
        \def\thiscolor{block1}
      \or
        \def\thiscolor{block2}
      \or
        \def\thiscolor{block3}
      \fi

      \filldraw[fill=\thiscolor, draw=black] (top-left) -- (top-right) -- (bottom-right) -- (bottom-left) -- cycle;
      \node at ($(g\gid-base)+(0,2.7)$) {};

      \foreach \dx/\dy in {0/0,1/0,0/1,1/1} {
        \pgfmathtruncatemacro{\xx}{\xH*2 + \dx}
        \pgfmathtruncatemacro{\yy}{\yH*2 + \dy}
        \coordinate (poly-bottom-\xx-\yy) at ($($(bottom-left)!0.5!(bottom-right)$) + (0,0)$);
        \draw[bend right=10, \thiscolor!90!black!50, thick] ($(g\gid-base)+(0,2.6)$) to ($(pix-\xx-\yy)+(0.1,0.1)$);
        }
     \filldraw[fill=\thiscolor, draw=black] (top-left) -- (top-right) -- (bottom-right) -- (bottom-left) -- cycle;
     \node at ($(g\gid-base)+(0,2.7)$) {[\minVal;\maxVal]};
    }
  }

\coordinate (top-center) at ($0.5*(g0-base)+0.5*(g3-base)+(4.5,4.5)$);
\coordinate (top-left) at ($(top-center)+(-0.45,0.45)$);
\coordinate (top-right) at ($(top-center)+(0.45,0.45)$);
\coordinate (bottom-right) at ($(top-center)+(0.45,-0.45)$);
\coordinate (bottom-left) at ($(top-center)+(-0.45,-0.45)$);
\node at (top-center) {$[1;16]$};
\foreach \gid in {0,1,2,3} {
    \ifcase\gid
        \def\thiscolor{block0}
      \or
        \def\thiscolor{block1}
      \or
        \def\thiscolor{block2}
      \or
        \def\thiscolor{block3}
    \fi
    \draw[bend left=10, \thiscolor!80!black!50, thick] (top-center) to ($(g\gid-base)+(0,2.4)$);
}
\filldraw[fill=white, draw=black] (top-left) -- (top-right) -- (bottom-right) -- (bottom-left) -- cycle;
\node at (top-center) {$[1;16]$};    

\end{scope}

\node[rotate=-25, right=6mm, above=6mm] at (pix-3-0) {Level 1};
\node[rotate=-25, left=12mm, above=6mm] at (g3-base) {Level 2};
\node[right=6mm, above=4mm, rotate=-25] at (top-center) {Level 3};

\end{tikzpicture}
    \subcaption{}\label{fig:quadtree-example}
  \end{subfigure}
  \vspace{0.5em} % optional spacing between subfigures
  \begin{subfigure}{\linewidth}
    \centering
    \definecolor{darkgreen}{rgb}{0.0, 0.5, 0.0}
\definecolor{block0}{RGB}{200,230,255}
\definecolor{block1}{RGB}{255,220,200}

\pgfdeclarelayer{bg}
\pgfsetlayers{bg,main}

\tikzset{iso/.style={x={(1.0cm,1.0cm)}, y={(-1.0cm,1.0cm)}}}

\begin{tikzpicture}[iso, scale=0.5, every node/.style={font=\tiny, transform shape}]
\node (1) at (-4,-4) [circle, draw, fill=block0, minimum size=0.8cm, inner sep=1pt] {$1$};
\node (2) at (-2,-4) [circle, draw, fill=block0, minimum size=0.8cm, inner sep=1pt] {$2$};
\node (3) at (0,-4)  [circle, draw, fill=block1, minimum size=0.8cm, inner sep=1pt] {$3$};
\node (4) at (2,-4)  [circle, draw, fill=block1, minimum size=0.8cm, inner sep=1pt] {$4$};

\node (12) at (-3,-2) [circle, draw, fill=block0] {$[1;2]$};
\node (34) at (1,-2)  [circle, draw, fill=block1] {$[3;4]$};
\node (1234) at (-1, 0) [circle, draw] {$[1;4]$};

\draw[block0!90!black!90, thick, bend right=25] (1) to (12);
\draw[block0!90!black!90, thick, bend left=25]  (2) to (12);
\draw[block1!90!black!90, thick, bend right=25] (3) to (34);
\draw[block1!90!black!90, thick, bend left=25]  (4) to (34);
\draw[block0!90!black!90, thick, bend right=45] (12) to (1234);
\draw[block1!90!black!90, thick, bend left=45]  (34) to (1234);

\node[draw, above right=0.025cm and 0.025cm of 1, minimum size=0.75cm, inner sep=0] (l1t) {It};
\node[draw, above right=0.025cm and 0.025cm of 2, minimum size=0.75cm, inner sep=0] (l2t) {is};
\node[draw, above right=0.025cm and 0.025cm of 3, minimum size=0.75cm, inner sep=0] (l3t) {a text};
\node[draw, above right=0.025cm and 0.025cm of 4, minimum size=0.75cm, inner sep=0] (l4t) {example};
\end{tikzpicture}
    \subcaption{}\label{fig:text-example}
  \end{subfigure}
  \caption{Constructing a tree graph for image and text inputs.}
  \vspace{-25pt}
  \label{fig:input_examples}
\end{wrapfigure}
\FloatBarrier
We consider a BFS-traverse ordered sequence $u_{1:L}$  as input and $x_{1:L}$ as output, with $k$-th element $x_k, u_k \in\mathbb{R}^{d_k}$. Hence, the first $1$ to $l_1$ elements constitute the first level of the tree, the next $l_1+1$ to $l_1+l_2+1$ the second level of the tree, and the last element with index $L$ is the root. We denote by $\pi$ a permutation mapping a BFS index to the post-order depth-first search (DFS) traverse. 

In this paper, we focus on text and image domains; hence we need to introduce a graph structure on top. A natural choice is a hierarchical structure. For the image domain, we consider a quad tree structure, where the first $l_1$ elements of the sequence correspond to the flattering of the z order (morton) of the image, and all the next levels are virtual nodes which cover sequentially larger segments of the image.  For the text domain, $n$-nary tree-cover chunks of tokens. In Figure \ref{fig:input_examples} we provide an illustration.

For classification tasks, there are two common strategies for information aggregation before MLP classification head. The first approach is to add a classification token at the end or in the middle of the sequence. In the Myo layer, the hidden state of the root node can serve as the state of the classification token naturally. Alternatively, one can average the hidden state over tokens. We propose to average the top $k$ layers of BFS (counting from the root), which interpolate between both common strategies, where $k=1$ corresponds to using only the hidden state of the root.

\paragraph{Layer parameters}
With each vertex of the tree $v\in V$, we associate three matrices $(A_v, B_v, C_v)$ as layer parameters. As the graph is a rooted tree, the association of $B_v$ and $C_v$ with the vertex $v$ is the same as the association with an edge $\{v, \partial^{*} v\}$. The matrices $B_v$, $C_v$ are responsible for the interaction between the vertex $v$ with its parent $\partial^{*}v$, and the matrix $A_v$ defines a self-interaction block. Using $\pi$ ordering of labels, we construct a block matrix $T_{G}$, where $A_v$ is a diagonal block at the index $(\pi(v), \pi(v))$ and off-diagonal blocks $B_v$ and $C_v$ are in the positions $(\pi(v), \pi(\partial^{*}v))$ and $(\pi(\partial^{*}v), \pi(v))$. In Figure \ref{fig:myostructure}, we provide an illustration of this process. 
The output of the Myo layer is the solution to the linear equation (with depth first post-order traverse ordering $\pi$ of matrix rows):
\begin{equation}
    \begin{aligned}
    \label{eq:solve}
    & T_G (\{A_v, B_v, C_v\}_{v=1}^{L}) \underbrace{x_{\pi}}_{\text{Output of the layer}} = \underbrace{u_{\pi}}_{\text{Input to the layer}}.  
    \end{aligned}
\end{equation}
For the general left part $T_G$ a direct solver has complexity $O(L^3)$, while an iterative solver will require several iterations with complexity $O(L^2)$ per iteration. We will take advantage of the tree structure and provide linear (in a single processor) and parallelized solutions. 

\subsection{Myo: Efficient Computation}
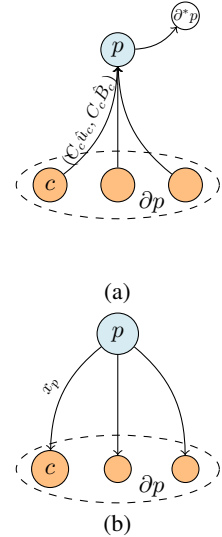
\begin{wrapfigure}[20]{r}{0.3\textwidth}
\centering
\begin{minipage}{0.28\textwidth}
    \centering
    \usetikzlibrary{decorations.text}
\usetikzlibrary{decorations.pathreplacing}
\definecolor{lightblue}{RGB}{173, 216, 230}
\begin{tikzpicture}[scale=0.9, transform shape]
\node (v) at (0,0) [circle, draw,  minimum size=0.5cm, inner sep=0, fill=lightblue, fill opacity=0.5, draw opacity=1., text opacity=1., draw=black] {$p$};

\node (c1) at (-1,-2) [circle, draw, minimum size=0.5cm, inner sep=0, fill=orange, fill opacity=0.5, draw opacity=1., text opacity=1., draw=black] {$c$};
\node (c2) at (0,-2) [circle, draw,  minimum size=0.5cm, inner sep=0, fill=orange, fill opacity=0.5, draw opacity=1., text opacity=1., draw=black]  {$~$};
\node (c3) at (1,-2) [circle, draw,  minimum size=0.5cm, inner sep=0, fill=orange, fill opacity=0.5, draw opacity=1., text opacity=1., draw=black]  {$~$};
\node (pv) at (1,0.5) [circle, draw,  minimum size=0.5cm, inner sep=0, scale=0.65]  {$\partial^{*}p$};

\draw [->, bend right=25] (c1) to node[pos=0.5, above, sloped, scale=0.75] {$(C_c\hat{u}_c, C_c\hat{B}_c)$} (v);
\draw[->] (c2) -- (v);
\draw[->, bend left=25] (c3) to (v);
\draw[->, bend right=25] (v) to (pv);

\draw[dashed] (0,-2.) ellipse (1.5 and 0.5);
\node at (0.5,-2.25) {$\partial p$};
\end{tikzpicture}
    \vskip -5pt
    \subcaption{}\label{fig:child2parent}
\end{minipage}
\begin{minipage}{0.28\textwidth}
    \centering
    \usetikzlibrary{decorations.text}
\usetikzlibrary{decorations.pathreplacing}
\definecolor{lightblue}{RGB}{173, 216, 230}
\begin{tikzpicture}[scale=0.9, transform shape]
\node (v) at (0,0) [circle, draw, fill=lightblue, fill opacity=0.5, draw opacity=1., text opacity=1., draw=black] {$p$};

\node (c1) at (-1,-2) [circle, draw, fill=orange, fill opacity=0.5, draw opacity=1., text opacity=1., draw=black] {$c$};
\node (c2) at (0,-2) [circle, draw, fill=orange, fill opacity=0.5, draw opacity=1., text opacity=1., draw=black]  {$~$};
\node (c3) at (1,-2) [circle, draw, fill=orange, fill opacity=0.5, draw opacity=1., text opacity=1., draw=black]  {$~$};

\draw [->, bend right=25] (v) to node[pos=0.5, above, sloped, scale=0.75] {$x_p$} (c1);
\draw[->] (v) -- (c2);
\draw[->, bend left=25] (v) to (c3);

\draw[dashed] (0,-2.) ellipse (1.5 and 0.5);
\node at (0.5,-2.25) {$\partial p$};
\end{tikzpicture}
    \vskip -25pt
    \subcaption{}\label{fig:parent2child}
\end{minipage}
\caption{\small{Message passing between parent and children nodes. A node $p$ (a) receives messages from  and 
(b) sends a message to children nodes $\partial p$.}}\label{fig:updown}
\end{wrapfigure}

Now we introduce an algorithm for solving Equation \ref{eq:solve} in $O(L)$ memory and $O(\log_{k} L)$ time complexity, when $G$ is $k$-nary perfect tree.
To build up the approach, we start with a 1-level rooted tree. We label leafs with $c\in\{1,\dots,k\}$ and the root $[1;k]$. We traverse from leaf to root (upward iteration) and backward (downward iteration), resembling Gaussian eliminations and substitutions steps. The upward iteration in block-matrix notation is the following:
\begin{equation}
\resizebox{0.75\textwidth}{!}{$
\left[
\begin{array}{c c c c|c}
    A_1      & O      &  O        & B_1       & u_1 \\
    \vdots   & \ddots &  \vdots   & \vdots    & \vdots \\
    O        & O      &  A_k      & B_k       &  u_k  \\  
    C_1      & \dots  &  C_k      & A_{[1;k]} & u_{[1;k]} 
\end{array}
\right]
\quad \Rightarrow \quad
\left[
\begin{array}{c c c c|c}
 I      & \dots    &  O        & A_1^{-1}B_1   & A_1^{-1}u_1 \\
 \vdots  & \ddots   &  \vdots   & \vdots        & \vdots \\
 O       & \dots    &  I        & A_k^{-1}B_k   & A_k^{-1}u_k  \\  
 O       & \dots    &  O        & A_{[1;k]} - \sum_{c=1}^{k} C_c A_c^{-1} B_c &
 u_{[1;k]} - \sum_{c=1}^{k} C_c A_c^{-1} u_c \\
\end{array}
\right]
$}
\end{equation}
We denote results in an upward iteration by putting a hat on a symbol: $\hat{B}_v = A_v^{-1}B_v$, $\hat{u}_v = A_v^{-1}u_v,~ \hat{A}_{v} = A_{v} - \sum_{c\in\partial v}C_c\hat{B}_c$.
We get the solution $x$ by the substitution from root to each leaf $c\in\{1,\dots, k\}$ (backward iteration):
\begin{equation}
\resizebox{0.95\textwidth}{!}{$
\begin{aligned}
  & \hat{A}_{[1;k]} = A_{[1;k]} - \sum_{c=1}^{k} C_c A_c^{-1}B_c, 
  & x_{[1;k]} = \hat{A}_{[1;k]}^{-1}\!\left(u_{[1;k]} - \sum_{c=1}^{k} C_c A_c^{-1}u_c\right),
  & x_c = A_c^{-1}u_c - (A^{-1}_cB_c)x_{[1;k]}.
\end{aligned}
$}
\end{equation}
From this example, we can see a general recursive algorithm. As in a rooted tree the path between any node and the root is unique, the recursion is well defined. Hence, we have an upward and backward traverse, illustrated in Figure \ref{fig:updown}.
\par\vspace{-\intextsep}
\paragraph{Upward traverse} 
In upward traverse, each vertex $v$ modifies its coefficients $A_v, B_v \to \hat{A}_v,\hat{B}_v$, its right-part $u_v \to \hat{u}_v$ and sends messages to its parent. Messages are aggregated across sibling nodes and update information in parent node: the on-diagonal block of coefficients and the right-part. Consider a parent $p$ and its children $c\in\partial p$, the updates are as follows:
\begin{equation}
    \begin{aligned}
    \hat{B}_c   &\leftarrow -A_c^{-1}B_c, &\hat{u}_c &\leftarrow A_c^{-1}u_c, \\
    \color{blue}{\hat{A}_{p}} &\leftarrow {\color{blue}{A_p}} + \underbrace{\sum_{c\in\partial p}\color{orange}{C_c\hat{B}_c,}}_{\text{Message from child nodes}} & \color{blue}{\hat{u}_p} &\leftarrow {\color{blue}{u_p}} - \underbrace{\sum_{c\in\partial p} \color{orange}{C_c\hat{u}_c}}_{\text{Message from child nodes}}.
    \end{aligned}
\end{equation}
As soon as a parent has received both messages, we consider it as a child of its own parent $\partial^{*}p$, and continue if the root has not been reached. 
\paragraph{Backward traverse} 
By construction, the upward traverse ends at the root level, with a single root node $R$ and block $\hat{A}_R x_R = \hat{u}_{R}$. We obtain the solution $x_R = \hat{A}_R^{-1}u_{R}$ as an initial condition and start recurrence from root to leaf over the unique path. Given a parent $p$ and its child nodes $c\in \partial^{*}p$, the solution in a child node is as follows:
\begin{equation}
    \begin{aligned}
        & {\color{orange}{x_c}} \leftarrow {\color{orange}{\hat{u}_c}} + \underbrace{\color{orange}{\hat{B}_c}\color{blue} {x_{\partial c}}}_{\text{Message from the parent node}}.
    \end{aligned}
\end{equation}
These computations are independent across equidistant from root children-parent groups. Hence, we can process them in parallel. To this end, we consider the breadth-first levels of the tree. Initializing children and parents as the first and second levels, we scan over tree across pairs of consecutive levels and apply computations described in upward and backward traverse. We illustrate this in Figure
\ref{fig:bfs_scan} and provide details and pseudocode in Appendix \ref{sec:app_pseudocode}. This leads to efficient parallel computation in $O(\log_n L)$ for a $n$-nary perfect tree.
\begin{figure}[h]
    \centering
    \usetikzlibrary{fit}
\definecolor{green}{rgb}{0.0, 0.5, 0.0}
\usetikzlibrary{shapes}
\pgfdeclarelayer{bg}
\pgfsetlayers{bg,main}
\usetikzlibrary{decorations.text}
\usetikzlibrary{decorations.pathreplacing}
\begin{adjustbox}{trim=0mm 8mm 0mm 0mm, clip}
\begin{tikzpicture}[scale=0.85, transform shape, every node/.style={minimum size=0.5cm, minimum width=0.5cm, align=center}]
\begin{scope}[scale=0.6, transform shape]
\begin{scope}[shift={(-4,0)}]
\node (1) at (-4,-4) [draw, minimum size=1cm, inner sep=1.5pt, fill=orange, fill opacity=0.25, draw opacity=1., text opacity=1., draw=black]  {$1$};
\node (2) at (-2,-4) [draw, minimum size=1cm, inner sep=1.5pt, fill=orange, fill opacity=0.25, draw opacity=1., text opacity=1., draw=black]  {$2$};
\node (3) at (0, -4) [draw, minimum size=1cm, inner sep=1.5pt, fill=orange, fill opacity=0.25, draw opacity=1., text opacity=1., draw=black]  {$3$};
\node (4) at (2, -4) [draw, minimum size=1cm, inner sep=1.5pt, fill=orange, fill opacity=0.25, draw opacity=1., text opacity=1., draw=black]  {$4$};

\node (5) at (4,-4) [draw, minimum size=1cm, inner sep=1.5pt, fill=orange, fill opacity=0.25, draw opacity=1., text opacity=1., draw=black]  {$5$};
\node (6) at (6,-4) [draw, minimum size=1cm, inner sep=1.5pt, fill=orange, fill opacity=0.25, draw opacity=1., text opacity=1., draw=black]  {$6$};
\node (7) at (8, -4) [draw, minimum size=1cm, inner sep=1.5pt, fill=orange, fill opacity=0.25, draw opacity=1., text opacity=1., draw=black]  {$7$};
\node (8) at (10, -4) [draw, minimum size=1cm, inner sep=1.5pt, fill=orange, fill opacity=0.25, draw opacity=1., text opacity=1., draw=black]  {$8$};

\node (12) at (-3,-2) [draw, minimum size=1cm, inner sep=1.5pt, fill=purple, fill opacity=0.25, draw opacity=1., text opacity=1., draw=black] {$[1;2]$};
\node (34) at (1,-2) [draw, minimum size=1cm, inner sep=1.5pt, fill=purple, fill opacity=0.25, draw opacity=1., text opacity=1., draw=black]  {$[3;4]$};
\node (1234) at (-1, 0) [draw, minimum size=1cm, inner sep=1.5pt, fill=green, fill opacity=0.25, draw opacity=1., text opacity=1., draw=black] {$[1;4]$};

\node (56) at (5,-2) [draw, minimum size=1cm, inner sep=1.5pt, fill=purple, fill opacity=0.25, draw opacity=1., text opacity=1., draw=black] {$[5;6]$};
\node (78) at (9,-2) [draw, minimum size=1cm, inner sep=1.5pt, fill=purple, fill opacity=0.25, draw opacity=1., text opacity=1., draw=black]  {$[7;8]$};
\node (5678) at (7, 0) [draw, minimum size=1cm, inner sep=1.5pt, fill=green, fill opacity=0.25, draw opacity=1., text opacity=1., draw=black] {$[5;8]$};

\node (12345678) at (3,2)  [draw, minimum size=1cm, inner sep=1.5pt, fill=blue, fill opacity=0.25, draw opacity=1., text opacity=1., draw=black] {$[1,8]$};

\draw[->, bend right=25] (1) to (12);
\draw[->, bend left=25] (2) to (12);
\draw[->, bend right=25] (3) to (34);
\draw[->, bend left=25] (4) to (34);
\draw[->, bend right=45] (12) to (1234);
\draw[->, bend left=45] (34) to (1234);

\draw[->, bend right=25] (5) to (56);
\draw[->, bend left=25] (6) to (56);
\draw[->, bend right=25] (7) to (78);
\draw[->, bend left=25] (8) to (78);
\draw[->, bend right=45] (56) to (5678);
\draw[->, bend left=45] (78) to (5678);

\draw[->, bend right=60] (1234) to (12345678);
\draw[->, bend left=60] (5678) to (12345678);
\end{scope}
\node (R11) at (11, -4) {$~$};
\node (R12) [right=of R11, xshift=-1.0cm, draw, minimum size=1cm, inner sep=1.5pt, fill=orange, fill opacity=0.35, draw opacity=1., text opacity=1., draw=black]  {$C_1$};
\node (R21) at (11, -2) {$f\Big($};
\node (R22) [right=of R21, xshift=-1.0cm, draw, minimum size=1cm, inner sep=1.5pt, fill=purple, fill opacity=0.35, draw opacity=1., text opacity=1., draw=black]  {$P_2$};
\node (R23) [right=of R22, xshift=-1.0cm,] {,};
\node (R24) [right=of R23, xshift=-1.0cm, draw, minimum size=1cm, inner sep=1.5pt, fill=orange, fill opacity=0.35, draw opacity=1., text opacity=1., draw=black]  {$C_1$};
\node (R25) [right=of R24, xshift=-1.0cm] {$\Big)=$};
\node (R26) [right=of R25, xshift=-1.0cm, draw, minimum size=1cm, inner sep=1.5pt, fill=yellow, fill opacity=0.35, draw opacity=1., text opacity=1., draw=black]  {$C_2$};

\node (R31) at (11, 0) {$f\Big($};
\node (R32) [right=of R31, xshift=-1.0cm, draw, minimum size=1cm, inner sep=1.5pt, fill=green, fill opacity=0.35, draw opacity=1., text opacity=1., draw=black]  {$P_3$};
\node (R33) [right=of R32, xshift=-1.0cm,] {,};
\node (R34) [right=of R33, xshift=-1.0cm, draw, minimum size=1cm, inner sep=1.5pt, fill=yellow, fill opacity=0.35, draw opacity=1., text opacity=1., draw=black]  {$C_2$};
\node (R35) [right=of R34, xshift=-1.0cm] {$\Big)=$};
\node (R36) [right=of R35, xshift=-1.0cm, draw, minimum size=1cm, inner sep=1.5pt, fill=yellow, fill opacity=0.35, draw opacity=1., text opacity=1., draw=black]  {$C_3$};

\node (R41) at (11, 2) {$f\Big($};
\node (R42) [right=of R41, xshift=-1.0cm, draw, minimum size=1cm, inner sep=1.5pt, fill=blue, fill opacity=0.35, draw opacity=1., text opacity=1., draw=black]  {$P_4$};
\node (R43) [right=of R42, xshift=-1.0cm,] {,};
\node (R44) [right=of R43, xshift=-1.0cm, draw, minimum size=1cm, inner sep=1.5pt, fill=yellow, fill opacity=0.35, draw opacity=1., text opacity=1., draw=black]  {$C_3$};
\node (R45) [right=of R44, xshift=-1.0cm] {$\Big)=$};
\node (R46) [right=of R45, xshift=-1.0cm, draw, minimum size=1cm, inner sep=1.5pt, fill=yellow, fill opacity=0.35, draw opacity=1., text opacity=1., draw=black]  {$C_4$};

\draw[->,  bend left=25] (R12) to (R24);
\draw[->,  bend left=25] (R26) to (R34);
\draw[->,  bend left=25] (R36) to (R44);
\end{scope}

\tikzset{
  wrap/.style 2 args={
    line cap=round,
    #1,
    line width=#2,
    opacity=0.10,
  },
}
\begin{pgfonlayer}{bg}
    \draw[wrap={orange}{5}](1.center) to[out=0,in=200](R12.center);
    \draw[wrap={orange}{5}](2.center) to[out=-5,in=200](R12.center);
    \draw[wrap={orange}{5}](3.center) to[out=-5,in=200](R12.center);
    \draw[wrap={orange}{5}](4.center) to[out=-5,in=200](R12.center);
    \draw[wrap={orange}{5}](5.center) to[out=-15,in=200](R12.center);
    \draw[wrap={orange}{5}](6.center) to[out=-15,in=200](R12.center);
    \draw[wrap={orange}{5}](7.center) to[out=-15,in=200](R12.center);
    \draw[wrap={orange}{5}](8.center) to[out=-15,in=200](R12.center);

    \draw[wrap={purple}{10}](12.center) to[out=0,in=210](R22.center);
    \draw[wrap={purple}{10}](34.center) to[out=-5,in=210](R22.center);
    \draw[wrap={purple}{10}](56.center) to[out=-10,in=210](R22.center);
    \draw[wrap={purple}{10}](78.center) to[out=-15,in=210](R22.center);

    \draw[wrap={green}{15}](1234.center) to[out=0,in=210](R32.center);
    \draw[wrap={green}{15}](5678.center) to[out=-15,in=210](R32.center);  

    \draw[wrap={blue}{20}](12345678.center) to[out=0,in=210](R42.center);   
\end{pgfonlayer}
%\draw (current bounding box.south east) rectangle (current bounding box.north west);
\end{tikzpicture}
\end{adjustbox}
    \captionsetup{skip=-12pt}
    \caption{Illustration of the upwards traverse over BFS levels of the tree. Levels data is stored in arrays. The function $f$ takes as input current value of carrying $C$ (children), and current level data as parent $P$ and output new $C'$, which is passed up.}
    \label{fig:bfs_scan}
\end{figure}
\subsection{Myo: SSM layer as particular case}
We show that a normal SSM layer is a particular case of the proposed layer, hence Myo is more general. 
Given a chain graph of length $L$, we consider the following system $T_c(\{(I, O, C_v)\}_{v=1}^{L})x_{1:L}=u_{1:L}$:
\begin{equation}
\resizebox{0.45\textwidth}{!}{$
\begin{bmatrix}
    I   & O     & O     &        &        \\
    C_2 & I     & O     &        &        \\
    O   & C_3   & I     & \ddots &        \\
        & \ddots& \ddots& \ddots & O      \\
        &       & O     & C_L    & I
\end{bmatrix}
\begin{bmatrix}
    x_{\pi(1)} \\[0.4em] \vdots \\[0.4em] x_{\pi(L)}
\end{bmatrix}
=
\begin{bmatrix}
    u_{\pi(1)} \\[0.4em] \vdots \\[0.4em] u_{\pi(L)}
\end{bmatrix}
$}
\end{equation}
\paragraph{Matching the SSM layer}
As a basis, consider a 3-chain with two edges $G_c(V,E): V_c=\{1,2,3\}, E_c=\{\{1,2\},\{2,3\}\}$. As the matrix is lower triangular, the solution can be obtained by direct substitution:
\begin{equation}
    \begin{aligned}
        \begin{bmatrix}
            I   & O   & O & | & u_1 \\
            C_2 & I   & O & | & u_2 \\
            O   & C_3 & I & | & u_3
        \end{bmatrix} \Rightarrow
        \begin{bmatrix}
            I   & O   & O & | & u_1 \\
            O   & I   & O & | & u_2 &-~C_2u_1  \\
            O   & O & I & | & u_3 &-~C_3u_2 &-~ C_3C_2u_1
        \end{bmatrix}.
    \end{aligned}
\end{equation}
Hence, since any chain graph with length $L>3$ is a union of overlapping 3-chains, the solution is the following recurrence:
\begin{equation}
    \begin{aligned}
        & x_1 = u_1, 
        & k\in\{2,\dots,L\}: x_k = - C_{k}x_{k-1} + u_k = u_k - \sum_{1\leq k < l} \left(
         \prod_{n=l}^{k+1}C_{n}\right) u_l.
    \end{aligned}
\footnote{$\prod_{k=1}^{3}a_k = a_1a_2a_3, \prod_{k=3}^{1}a_k = a_3a_2a_1$}
\end{equation}
It follows that the inverse of the lower bidiagonal matrix $T_c$ is a dense low-triangular matrix. The value of $(T^{-1}_c)_{ij}$ is the $i$-th coordinate of the solution $T_c x_{1:L} = u_{1:L}$ with the right part $u = (O,\dots \underbrace{I}_{j\text{-th position}}, \dots, O)$:
\begin{equation}
    \begin{aligned}
    (T^{-1}_c)_{ij} = 
    \left\{\begin{matrix}
    & O + \delta_{ij} I, & \text{if } i \geq j, \\
    & -\prod_{n=j}^{i+1}C_{n} & \text{if } i < j.
    \end{matrix}\right.
    \end{aligned}
\end{equation}
To match the SSM layer, we need to also add a projection of the right part, for any left part $T_G$ this is just a lock-diagonal rescaling with $A_k$. Recalling the definition of the SSM layer:
\begin{equation}
    \begin{aligned}
    & x_{1} = S_1u_1, & k\in\{2,\dots, L\}:
    x_k = I_{k-1}x_{k-1} + S_{k}u_k.
    \end{aligned}
\end{equation}
Hence, we have following correspondence:
\begin{table}[h]
\centering 
\begin{tabular}{|c|c|c|}
\hline
& SSM & Myo on the Chain graph \\
\hline
Self-term (named $S$ in SSM, $A$ in Myo) & $S:B_k$ & $S:B_k^{-1}$ \\
Interaction term (named $I$ in SSM, $B,C$ in Myo) & $I:A_k$ & $I (B_k,C_k): (O, -B_k^{-1}A_k)$ \\
\hline
\end{tabular}
\label{tab:ssm2myo}
\caption{Parameters correspondence for Myo on Chain graph in order to match given SSM layer.}
\end{table}
\paragraph{Bidirectional SSM on a chain graph}
We note that an SSM layer has a preferable ordering, hence the $i$ th element of the output sequence is influenced only by its predecessors and the information in elements after $i$ is ignored. To mitigate this, a common approach is to reverse the input sequence $u_{1:L}$ and apply the same layer to both $u_{1:L}$ and $u_{L:1}$, stacking the output as channels. In contrast, Myo uses two interaction matrices $(C_v, B_v)$ per edge, where $B_v$ corresponds to the interaction of the vertex $v$ with the parent $\partial^{*} v$ and $C_v$ vice versa. In this case, the chain graph corresponds to the tri-diagonal matrix. Any tridiagonal matrix can be represented as the composition of lower bi-diagonal and upper bi-diagonal matrices. Hence, the application of a Myo layer on a chain graph corresponds to two consecutive applications of an SSM: first to the input $u_{1:L}$ and second to the reversed result.

\paragraph{Why beyond chains?}
Although the s chain is a tree, it is a very limited one, as any parent has exactly one child. As we shall see next, having several children leads to more interesting aggregation of information between nodes. Also, the tree structure allows us to more naturally map the neighborhoods, than flattening.

\subsection{Myo: Parametrization}
\paragraph{Partial Gauge Fixing} 
Consider again a children-parent block:
$
    \begin{bmatrix}
        A_1 & O   &  O   & B_1 \\
        \vdots   & \ddots &  \vdots  & \vdots \\
        O   & O   &  A_k & B_k  \\  
        C_1 & \dots &  C_k & A_{[1;k]}
    \end{bmatrix}
$.
Transformation of the input vector $u_{1:L}$ with a block diagonal matrix $D_v$ has the same effect on the output of the layer, as changing the layer parameters as follows: $A_v \to D^{-1}_vA_v, B_v\to D^{-1}_{\partial v}B_v$. We can partially fix redundancy by fixing all $A_i$ equal to $I$. Note that it does not imply that diagonal elements will act trivially, i.e. diagonal blocks of inverted matrix are not identity. In order to show this, consider diagonal blocks during the upward pass:
\begin{equation}
\resizebox{0.98\textwidth}{!}{$
\begin{array}{ccc}
\text{for first (leaf) level node \(c\)} &
\text{for second level node \(v\)} &
\text{for third level node \(w\)} \\
\hat{A}_{c} = I, &
\hat{A}_{v} = I + \sum_{c\in\partial^{*}v}C_cB_c, &
\hat{A}_{w} = I + \sum_{v\in\partial^{*}w}C_v
\left(I + \sum_{g\in\partial^{*}v}C_gB_g\right)^{-1}B_v
\end{array}
$}
\end{equation}
For stability of training, we consider the matrix to be diagonally dominant and symmetric $C_v^{\dag} = B_v$. 
We initialize the transition matrix blocks $B_i$ as \cite{smith2023simplified}, diagonalizing the HIPPO-N matrix,
\section{Experiments}
For all experiments, we take the architecture from \cite{smith2023simplified} and only change the SSM block to the Myo. The architecture consists of linear encoder, stacks of Myo layers with skip connection, and silu nonlinearity. For fairness of comparison, we did not use any augmentations, following a common experiment design on Long-Range Arena datasets. For all experiments, the block size was set to 1, that is, scalar, and the number of heads was selected to match the state dimension of the S4 and S5 models. All experiments were performed in a single NVIDIA V100 GPU accelerator.
\paragraph{Pixel-level 1d image classification}
We report results on classification tasks, including sequential CIFAR (3 channels) and sequential MNIST benchmarks. Both datasets flattened in common sequential versions: sMNIST, sCIFAR with snake order, and in quadtree aligned versions: zMNIST, zCIFAR wth morton ordering, see Figure \ref{fig:morton-snake}. In Appendix \ref{sec:datasets} we provide a description of the datasets and flattering procedure. For consistency, we used perfect trees with four children in Myo layers in both tasks. See Table~\ref{tab:pixellong_short} for results and the full Table~\ref{tab:pixellong} in Appendix~\ref{sec:app_fullt} (we omit some non-top scores for space considerations). 
\begin{table}[t!]
  \captionsetup{justification=centering}
  \caption{Test accuracy on image classification. We use the table from \cite{smith2023simplified} and add our results.}
  \label{tab:pixellong_short}
  \centering
  \begin{adjustbox}{center}
    \begin{tabular}{@{}lcccc@{}}
    \toprule
    Model               & \texttt{sMNIST}  & \texttt{sCIFAR} & \texttt{zMNIST} & \texttt{zCIFAR} \\ 
    (Input length)      & (784)            & (1024)          & (784)           & (1024)          \\ \midrule    
    Transformer \citep{trinh2018learning, vaswani2017attention} & 98.9  & 62.2   & - & - \\  \midrule 
    CCNN \citep{romero2022towards}         & \textbf{99.72}  & \textbf{93.08}  & - & - \\
    LSTM  \citep{gu2020improving, hochreiter1997long} & 98.9 & 63.01  & - & - \\
    r-LSTM   \citep{trinh2018learning}     & 98.4            & 72.2            & - & - \\
    HiPPO-RNN  \citep{gu2020hippo}         & 98.9            & 61.1            & - & - \\
    S4    \citep{gu2022parameterization, gu2021s4} & 99.63     & 91.80           & - & - \\
    S4D   \citep{gu2022parameterization}   & -               & 89.92           & - & - \\
    Liquid-S4 \citep{hasani2022liquid}     & -               & 92.02           & - & - \\
    S5                                       & 99.65         & 90.10           & 99.5 (our run) & 89.9 (our run) \\ \bottomrule
    \textbf{Myo} (this work) & 99.2 & 92.6 & \textbf{99.7} & \textbf{93.4} \\ \bottomrule
    \end{tabular}
\end{adjustbox}
\end{table}
The results in Table \ref{tab:pixellong_short} suggest that without tree structure-aware flattering of an image, Myo performs on par with other architectures. However, if Morton ordering is used, which is aware of the quad tree structure, the results are slightly better. We were able to run only S5 model on Morton flattering, however, we do not expect that other architectures will improve their results with changing flattering order. 
\paragraph{Subset of Long Range Arena benchmark}
We consider a binary classification task of flattened images (we keep the snake order, as a benchmark introduced) of PathX dataset and a 10-way classification task in Listops benchmark. See Appendix \ref{sec:datasets} for a detailed description of the tasks. We consider both tasks as given text sequences and use a binary perfect tree. We present results in Table \ref{tab:lra_simplified}. Without tree-structured aware flattering, Myo performs comparable, but not better. 
\begin{table}
  \captionsetup{justification=centering}
  \caption{Test accuracy on selected LRA benchmark tasks. \xmark\ indicates the model did not exceed random guessing. We used the table from \cite{smith2023simplified} and add our results.}
  \label{tab:lra_simplified}
  \centering
  \begin{adjustbox}{center}
    \begin{tabular}{@{}lccc@{}}
    \toprule
    Model               & \texttt{ListOps} & \texttt{Pathfinder} & \texttt{Path-X} \\
    (Input length)      & (2048)          & (1024)            & (16384)        \\ \midrule
    Transformer         & 36.37            & 71.40              & \xmark          \\
    Luna-256            & 37.25            & 77.72              & \xmark          \\
    H-Trans.-1D         & 49.53            & 68.78              & \xmark          \\
    CCNN                & 43.60            & 91.51              & \xmark          \\ \midrule
    Mega ($\mathcal{O}(L^2)$)  & \textbf{63.14} & \textbf{96.01} & 97.98 \\
    Mega-chunk ($\mathcal{O}(L)$) & 58.76       & 94.41          & 93.81          \\ \midrule
    S4D-LegS            & 60.47            & 93.06              & 91.95          \\
    S4-LegS             & 59.60            & 94.20              & 96.35          \\
    Liquid-S4           & 62.75 & 94.80             & 96.66          \\ 
    S5         & 62.15            & 95.33  & \textbf{98.58} \\ \midrule
    \textbf{Myo} (this work)                 & 59.5              & 86.1                 & 85.7           \\ \bottomrule
    \end{tabular}
  \end{adjustbox}
\end{table}
\section{Conclusion}
We present Myosotis, an SSM and attention-like layer based on an efficient recurrent inversion of the quad-tree-structured matrix. The benchmarks suggest that when the data align with the quad-tree structure, Myosotis is a superior choice, otherwise performing on par with SSM. We believe that this opens up new possibilities on a structured layer design. 

\newpage
\bibliographystyle{unsrt}
\bibliography{bib.bib}

\begin{thebibliography}{10}

\bibitem{vaswani2017attention}
Ashish Vaswani, Noam Shazeer, Niki Parmar, Jakob Uszkoreit, Llion Jones, Aidan~N Gomez, {\L}ukasz Kaiser, and Illia Polosukhin.
\newblock Attention is all you need.
\newblock {\em Advances in Neural Information Processing Systems}, 30, 2017.

\bibitem{zaheer2020bigbird}
Manzil Zaheer, Guru Guruganesh, Kumar~Avinava Dubey, Joshua Ainslie, Chris Alberti, Santiago Ontanon, Philip Pham, Anirudh Ravula, Qifan Wang, Li~Yang, et~al.
\newblock Big bird: Transformers for longer sequences.
\newblock {\em Advances in Neural Information Processing Systems}, 33, 2020.

\bibitem{beltagy2020longformer}
Iz~Beltagy, Matthew Peters, and Arman Cohan.
\newblock Longformer: The long-document transformer.
\newblock {\em arXiv preprint arXiv:2004.05150}, 2020.

\bibitem{choromanski2021rethinking}
Krzysztof~Marcin Choromanski, Valerii Likhosherstov, David Dohan, Xingyou Song, Andreea Gane, Tamas Sarlos, Peter Hawkins, Jared~Quincy Davis, Afroz Mohiuddin, Lukasz Kaiser, David~Benjamin Belanger, Lucy Colwell, and Adrian Weller.
\newblock Rethinking attention with performers.
\newblock In {\em International Conference on Learning Representations}, 2021.

\bibitem{liu2021Swin}
Ze~Liu, Yutong Lin, Yue Cao, Han Hu, Yixuan Wei, Zheng Zhang, Stephen Lin, and Baining Guo.
\newblock Swin transformer: Hierarchical vision transformer using shifted windows.
\newblock In {\em Proceedings of the IEEE/CVF International Conference on Computer Vision (ICCV)}, 2021.

\bibitem{gu2020hippo}
Albert Gu, Tri Dao, Stefano Ermon, Atri Rudra, and Christopher R{\'e}.
\newblock Hippo: Recurrent memory with optimal polynomial projections.
\newblock {\em Advances in Neural Information Processing Systems}, 33, 2020.

\bibitem{gu2021combining}
Albert Gu, Isys Johnson, Karan Goel, Khaled Saab, Tri Dao, Atri Rudra, and Christopher R{\'e}.
\newblock Combining recurrent, convolutional, and continuous-time models with linear state-space layers.
\newblock {\em Advances in Neural Information Processing Systems}, 34, 2021.

\bibitem{gu2022efficiently}
Albert Gu, Karan Goel, and Christopher R\'e.
\newblock Efficiently modeling long sequences with structured state spaces.
\newblock In {\em The International Conference on Learning Representations ({ICLR})}, 2022.

\bibitem{gu2023hippo}
Albert Gu, Isys Johnson, Aman Timalsina, Atri Rudra, and Christopher R\'e.
\newblock How to train your hippo: State space models with generalized basis projections.
\newblock In {\em The International Conference on Learning Representations ({ICLR})}, 2023.

\bibitem{nguyen2022s4nd}
Eric Nguyen, Karan Goel, Albert Gu, Gordon~W. Downs, Preey Shah, Tri Dao, Stephen~A. Baccus, and Christopher R\'e.
\newblock S4nd: Modeling images and videos as multidimensional signals using state spaces.
\newblock {\em Advances in Neural Information Processing Systems}, 35, 2022.

\bibitem{goel2022sashimi}
Karan Goel, Albert Gu, Chris Donahue, and Christopher R{\'e}.
\newblock It's raw! audio generation with state-space models.
\newblock {\em International Conference on Machine Learning ({ICML})}, 2022.

\bibitem{gu2021s4}
Albert Gu, Karan Goel, and Christopher Re.
\newblock Efficiently modeling long sequences with structured state spaces.
\newblock In {\em International Conference on Learning Representations}, 2021.

\bibitem{smith2023simplified}
Jimmy~T.H. Smith, Andrew Warrington, and Scott Linderman.
\newblock Simplified state space layers for sequence modeling.
\newblock In {\em The Eleventh International Conference on Learning Representations}, 2023.

\bibitem{gu2022s4d}
Albert Gu, Ankit Gupta, Karan Goel, and Christopher R\'e.
\newblock On the parameterization and initialization of diagonal state space models.
\newblock {\em Advances in Neural Information Processing Systems}, 35, 2022.

\bibitem{trinh2018learning}
Trieu Trinh, Andrew Dai, Thang Luong, and Quoc Le.
\newblock Learning longer-term dependencies in {RNN}s with auxiliary losses.
\newblock In {\em International Conference on Machine Learning}, pages 4965--4974. PMLR, 2018.

\bibitem{romero2022towards}
David Romero, David Knigge, Albert Gu, Erik Bekkers, Efstratios Gavves, Jakub Tomczak, and Mark Hoogendoorn.
\newblock Towards a general purpose {CNN} for long range dependencies in {$ND$}.
\newblock {\em arXiv preprint arXiv:2206.03398}, 2022.

\bibitem{gu2020improving}
Albert Gu, Caglar Gulcehre, Thomas Paine, Matt Hoffman, and Razvan Pascanu.
\newblock Improving the gating mechanism of recurrent neural networks.
\newblock In {\em International Conference on Machine Learning}, pages 3800--3809. PMLR, 2020.

\bibitem{hochreiter1997long}
Sepp Hochreiter and J{\"u}rgen Schmidhuber.
\newblock Long short-term memory.
\newblock {\em Neural {C}omputation}, 9(8):1735--1780, 1997.

\bibitem{gu2022parameterization}
Albert Gu, Karan Goel, Ankit Gupta, and Christopher R{\'e}.
\newblock On the parameterization and initialization of diagonal state space models.
\newblock In {\em Advances in Neural Information Processing Systems}, 2022.

\bibitem{hasani2022liquid}
Ramin Hasani, Mathias Lechner, Tsun-Hsuan Wang, Makram Chahine, Alexander Amini, and Daniela Rus.
\newblock Liquid structural state-space models.
\newblock In {\em International Conference on Learning Representations}, 2023.

\bibitem{krizhevsky2009learning}
Alex Krizhevsky.
\newblock Learning multiple layers of features from tiny images.
\newblock {\em Master's thesis, University of Toronto}, 2009.

\bibitem{tay2020lra}
Yi~Tay, Mostafa Dehghani, Samira Abnar, Yikang Shen, Dara Bahri, Philip Pham, Jinfeng Rao, Liu Yang, Sebastian Ruder, and Donald Metzler.
\newblock Long {R}ange {A}rena: A benchmark for efficient transformers.
\newblock In {\em International Conference on Learning Representations}, 2021.

\bibitem{linsley2018learning}
Drew Linsley, Junkyung Kim, Vijay Veerabadran, Charles Windolf, and Thomas Serre.
\newblock Learning long-range spatial dependencies with horizontal gated recurrent units.
\newblock {\em Advances in Neural Information Processing Systems}, 31, 2018.

\bibitem{nangia2018listops}
Nikita Nangia and Samuel Bowman.
\newblock {ListOps}: A diagnostic dataset for latent tree learning.
\newblock {\em NAACL HLT 2018}, page~92, 2018.

\bibitem{romero2021flexconv}
David Romero, Robert-Jan Bruintjes, Jakub~Mikolaj Tomczak, Erik Bekkers, Mark Hoogendoorn, and Jan van Gemert.
\newblock Flexconv: Continuous kernel convolutions with differentiable kernel sizes.
\newblock In {\em International Conference on Learning Representations}, 2021.

\bibitem{romero2021ckconv}
David Romero, Anna Kuzina, Erik Bekkers, Jakub~Mikolaj Tomczak, and Mark Hoogendoorn.
\newblock {CKC}onv: Continuous kernel convolution for sequential data.
\newblock In {\em International Conference on Learning Representations}, 2022.

\bibitem{bai2018trellis}
Shaojie Bai, J.~Zico Kolter, and Vladlen Koltun.
\newblock Trellis networks for sequence modeling.
\newblock In {\em International Conference on Learning Representations}, 2019.

\bibitem{bai2018empirical}
Shaojie Bai, J.~Zico Kolter, and Vladlen Koltun.
\newblock An empirical evaluation of generic convolutional and recurrent networks for sequence modeling.
\newblock {\em arXiv preprint arXiv:1803.01271}, 2018.

\bibitem{chang2017dilated}
Shiyu Chang, Yang Zhang, Wei Han, Mo~Yu, Xiaoxiao Guo, Wei Tan, Xiaodong Cui, Michael Witbrock, Mark~A Hasegawa-Johnson, and Thomas~S Huang.
\newblock Dilated recurrent neural networks.
\newblock {\em Advances in Neural Information Processing Systems}, 30, 2017.

\bibitem{li2018independently}
Shuai Li, Wanqing Li, Chris Cook, Ce~Zhu, and Yanbo Gao.
\newblock Independently recurrent neural network ({INDRNN}): Building a longer and deeper {RNN}.
\newblock In {\em Proceedings of the IEEE Conference on Computer Vision and Pattern Recognition}, pages 5457--5466, 2018.

\bibitem{lezcano2019cheap}
Mario Lezcano-Casado and David Mart{\i}nez-Rubio.
\newblock Cheap orthogonal constraints in neural networks: A simple parametrization of the orthogonal and unitary group.
\newblock In {\em International Conference on Machine Learning}, pages 3794--3803. PMLR, 2019.

\bibitem{Voelker2019LegendreMU}
Aaron Voelker, Ivana Kaji{\'c}, and Chris Eliasmith.
\newblock Legendre {M}emory {U}nits: Continuous-time representation in recurrent neural networks.
\newblock {\em Advances in Neural Information Processing Systems}, 32, 2019.

\bibitem{rusch2021unicornn}
T.~Konstantin Rusch and Siddhartha Mishra.
\newblock Unicornn: A recurrent model for learning very long time dependencies.
\newblock In {\em International Conference on Machine Learning}, pages 9168--9178. PMLR, 2021.

\bibitem{chilkuri2021parallelizing}
Narsimha~Reddy Chilkuri and Chris Eliasmith.
\newblock Parallelizing {L}egendre memory unit training.
\newblock In {\em International Conference on Machine Learning}, pages 1898--1907. PMLR, 2021.

\bibitem{erichson2021lipschitz}
N.~Benjamin Erichson, Omri Azencot, Alejandro Queiruga, Liam Hodgkinson, and Michael Mahoney.
\newblock Lipschitz recurrent neural networks.
\newblock In {\em International Conference on Learning Representations}, 2021.

\bibitem{gu2021lssl}
Albert Gu, Isys Johnson, Karan Goel, Khaled Saab, Tri Dao, Atri Rudra, and Christopher R{\'e}.
\newblock Combining recurrent, convolutional, and continuous-time models with linear state space layers.
\newblock {\em Advances in Neural Information Processing Systems}, 34, 2021.

\end{thebibliography}
\newpage

\appendix
\section{Appendix: Details of forward pass computation}\label{sec:app_pseudocode}
\subsection{Myo: Shape Description}
In this section, we elaborate on the tensor shapes in the Myo layer. For concreteness, we consider a basic block: a parent that covers the $k$ child nodes, at some level $l$ (the parent node is at level $l+1$)
\begin{equation}
\begin{aligned}
\begin{bmatrix}
        A^{l}_{1} & O   &  O   & B^{l}_1 &       | & u^{l}_1 \\
        \vdots   & \ddots &  \vdots   & \vdots &       | & \vdots \\
        O   & O   &  A^{l}_k & B^{l}_k &       | & u^{l}_k  \\  
        C^{l}_1 & \dots &  C^{l}_k & A^{l+1}_{[1;k]} & | & u^{l+1}_{[1;k]} \\
    \end{bmatrix}.
\end{aligned}
\end{equation}
The choice of dimension of diagonal blocks fixes measurements of off-diagonal blocks. Given the input $u^{l}$ with dimension $D_{l} = r\times d_{l}$, the size of the block can be equal to $(D_{l},D_{l})$ or apply a layer with block size $(d_{l}, d_{l})$ to $r$ different vectors $u$. On top of that, the layer can have a head dimension $H$, i.e. application of different parameters to the same input.
Hence, the parameters of the layer $\mathcal{A}$ are given by a length list $d$ (for each BFS layer), where each element $A_l$ has dimensions $(H, L_l, d_l, d_l)$ and input $\mathcal{U}$, with each element $u_{l}$ with dimensions $(H, L_l, d_l, r)$ (one can add as many leading dimensions as desired, for example batch). Next we describe the algorithm in pseudo-code both abstract and python-style.

\subsection{Pseudo-code}
Here we provide an abstract pseudocode without batch and head dimensions; see the next section \ref{sec:python_app_pseudocode} for elaboration on them. We consider a tree $T$ with $L$ BFS layers, where $1$ indicates the level of the leaf and $L$ the level of the root. Let $\mathcal{A}_{1:L}, \mathcal{B}_{1:L}, \mathcal{C}_{1:L}, \mathcal{U}_{1:L}$ be tuples of length $L$. For each $l$ in $\mathcal{A}_{1:L}$, 
$\text{shape}(A_l)$ is $(n_{l}, d_{l}, d_{l})$, where $n_{l}$ is the number of nodes at the level $l$, and $d_l$ is the size of the block. Then $\text{shape}(B_l)$ is $(n_{l}, d_{l}, d_{l+1})$ and $\text{shape}(B_l)$ is $(n_{l}, d_{l+1}, d_{l})$. Finally, $\text{shape}(u_l)$ is $(n_{l}, d_{l}, r)$, where $r$ is a number of right parts, that is, we solve $T[x^1_{1:L}, \dots, x^R_{1:L}]=[u^1_{1:L}, \dots, u^R_{1:L}]$. We enumerate dimensions as $(1,2,3)$ and refer to them as numbers in pseudo-code where the function is applicable and as ":", where the function is vectorized.
We use \textit{split} to denote splitting of the layer $l$ array into chunks, where each chunk corresponds to the parent-child array at level $l+1$. If a tree is complete, then the BFS ordering guarantees that this holds; otherwise, it is always possible to sort branches not from the first level at most right of the tree.
\captionsetup[algorithm]{font={small}, labelfont={small,bf}}
\renewcommand{\alglinenumber}[1]{\scriptsize #1}
\noindent
\begin{minipage}[t]{0.48\textwidth}
\vtop{
\begin{algorithm}[H]
\caption{Upward Traverse}
\scriptsize
\begin{algorithmic}[1]
\Procedure{UpF}{$\{A_k, B_k, C_k, u_k\}_{k\in\{c,p\}}$}
    \State $\hat{B}_c \leftarrow - \text{Solve}_{:,2,3}(A_c, B_c)$
    \State $\hat{u}_c \leftarrow \text{Solve}_{:,2,:}(A_c, u_c)$
    \For{each $(b_{c,p}, c_{c,p})$ in split $(\hat{B}_c, C_c)$}
        \State $\hat{A}_{p, \partial^{*}c} \leftarrow A_{p, \partial^{*}c} + \text{einsum}(lik, ljk)(c_{c,p}, b_{c,p})$
        \State $\hat{u}_{p, \partial^{*}c} \leftarrow u_{p, \partial^{*}c} - \text{einsum}(lik, ljk)(c_{c,p}, \hat{u}_{c,p})$
    \EndFor
    \State \Return $(\hat{A}_p, B_p, C_p, \hat{u}_p), (\hat{u}_c, \hat{B}_c)$
\EndProcedure
\end{algorithmic}
\end{algorithm}

\begin{algorithm}[H]
\caption{Downward Traverse}
\scriptsize
\begin{algorithmic}[1]
\Procedure{SolveChP}{$B_c, \{u_k\}_{k\in\{c,p\}}$}
    \For{each $(b_{c,p}, u_{p})$ in split $(B_c, u_p)$}
        \State $b_{c} \leftarrow \text{einsum(cik, kj)}({b_{c,p}, u_{p}})$
    \EndFor
    \State $u_c \leftarrow u_c + b_c$
    \State \Return $u_c$
\EndProcedure
\end{algorithmic}
\end{algorithm}
}
\end{minipage}
\hfill
\begin{minipage}[t]{0.48\textwidth}
\vtop{
\begin{algorithm}[H]
\caption{Forward pass}
\scriptsize
\begin{algorithmic}[1]
\Procedure{Forward pass}{$\mathcal{A}_{1:L}, \mathcal{B}_{1:L}, \mathcal{C}_{1:L}, \mathcal{U}_{1:L}$}
    \State $A_c, B_c, C_c, u_c \leftarrow \mathcal{A}_{1}, \mathcal{B}_{1},\mathcal{C}_{1}, \mathcal{U}_{1}$
    \State $\mathcal{I}_{1:L}, \mathcal{X}_{1:L} \leftarrow \text{List}\{\text{empty}\}, \text{List}\{\text{empty}\}$
    \For{each $p$ in $2:L$}
        \State $(A_c, B_c, C_c, u_c), \mathcal{I}_{1:L}[p] \leftarrow$
        \Statex \hspace{1em} $\text{UpF}(\{A_k, B_k, C_k, u_k\}_{k\in\{c,p\}})$
    \EndFor
    \State $\mathcal{X}_{1:L}[1] \leftarrow \text{Solve}_{:,2:,}(A_{c}, u_{c})$
    \State $x \leftarrow \mathcal{X}_{1:L}[1]$
    \For{each $u, b$ in $\text{reversed}(\mathcal{I}_{1:L})$}
        \State $x \leftarrow \text{SolveChP}(u, b, x)$
        \State $\mathcal{X}_{1:L}.\text{appendleft}(x)$
    \EndFor
    \State\Return $\mathcal{X}_{1:L}$
\EndProcedure
\end{algorithmic}
\end{algorithm}
}
\end{minipage}
\par\vspace{-\intextsep}
\vskip -50pt
\newpage
\subsection{Python style pseudo-code}\label{sec:python_app_pseudocode}
Here, "b" is a batch dimension, "h" is a head dimension, "p", "c" is a dimension corresponding to the nodes over the layer, and the last two dimensions are either block dimensions for coefficients or block dimension and number of right parts for right part input.  
\begin{minipage}{0.99\textwidth}
    \begin{center}
    \begin{lstlisting}[language=Python, basicstyle=\scriptsize\ttfamily, showlines=true, caption=Python style implementation]
    
    CarryLeaf2Root: TypeAlias = Tuple[Float[Array, "b h p m m"], Float[Array, "b h p m k"], 
                                      Float[Array, "b h p k m"], Float[Array, "b h p m r"], 
                                      Tuple[int, ...]]
    
    YLeaf2Root: TypeAlias = Tuple[Float[Array, "b h c m r"], 
                                  Float[Array, "b h c m k"], 
                                  Tuple[int, ...]]
        
    def solve_leaf2root_scan_f(carry: CarryLeaf2Root,
                               x: CarryLeaf2Root) -> Tuple[CarryLeaf2Root, YLeaf2Root]:
        # carry
        # a b c y num
        # 0 1 2 3 4
        Ap, Bp, Cp, Yp, nump = x
    
        B = update_B_leafs(carry[0], carry[1])
        Yl = update_right_part_leaf(carry[0], carry[3])
        MC = make_leaf2parent_coeff_message(carry[2], B, carry[4])
        A = sum_update_coeff_parent(Ap, MC)
        MY = make_leaf2parent_rp_message(carry[2], Yl, carry[4])
        Y = update_right_part_parent(Yp, MY)
    
        return (A, Bp, Cp, Y, nump), (Yl, B, carry[4])

    def solve_leaf_given_parent(Yl: Float[Array, "b h c m r"],
                                B: Float[Array, "b h c m n"],
                                Yp: Float[Array, "b h p n r"],
                                split: Tuple[int, ...]) -> Float[Array, "b h c m r"]:
        b = split(B, split, axis=2)
        Ypp = split(Yp, axis=2)
        b = concatenate([einsum('...cik, ...kj -> ...cij', bb, yy) for bb, yy in zip(b, Ypp)], axis=2)
        return Yl + b
    
    def solve_schur_scan(a: List[Float[Array, "b h l m m"]], 
                        b: List[Float[Array, "b h l m k"]], 
                        c: List[Float[Array, "b h l k m"]], 
                        y: List[Float[Array, "b h l m r"]], 
                        split: Tuple[Tuple[int, ...], ...]) -> List[Float[Array, "b h l m r"]]:
    
        parents = zip(a[1:], b[1:], c[1:], y[1:], split[1:])
        carry = a[0], b[0], c[0], y[0], split[0]
        ys = []
    
        for p in parents:
            carry, y = solve_leaf2root_scan_f(carry, p)
            ys.append(y)
    
        yr = deque([dense_solve(carry[0], carry[3])])
        carry = yr[-1]
        
        for Yc, Bc, numc in reversed(ys):
            carry = solve_leaf_given_parent(Yc, Bc, carry, numc)
            yr.appendleft(carry)
    
        return list(yr)
    
    
    \end{lstlisting}
    \end{center}
    \end{minipage}

\subsection{Datasets description}\label{sec:datasets}
\begin{figure}[h]
\centering
\begin{tikzpicture}[scale=0.8, every node/.style={font=\small}]
\begin{scope}[shift={(0,0)}]
  \foreach \x in {0,...,3} {
    \foreach \y in {0,...,3} {
      \pgfmathtruncatemacro{\blockid}{floor(\x/2) + 2*floor(\y/2)}
      \ifcase\blockid
        \def\thiscolor{red!20}
      \or
        \def\thiscolor{green!20}
      \or
        \def\thiscolor{blue!20}
      \or
        \def\thiscolor{yellow!20}
      \fi

      \filldraw[fill=\thiscolor, draw=black] (\x,\y) rectangle ++(1,1);
      \coordinate (m-\x-\y) at (\x+0.5,\y+0.5);

      \pgfmathtruncatemacro{\xL}{mod(\x,2)}
      \pgfmathtruncatemacro{\yL}{mod(\y,2)}
      \pgfmathtruncatemacro{\xH}{floor(\x/2)}
      \pgfmathtruncatemacro{\yH}{floor(\y/2)}
      \pgfmathtruncatemacro{\morton}{\xL + 2*\yL + 4*\xH + 8*\yH + 1}

      \node (m\morton) at (m-\x-\y) {\morton};
    }
  }

  \foreach \i [evaluate=\i as \j using int(\i+1)] in {1,...,15} {
    \ifodd\i
        \draw[->, bend right=2, red!90!black, thin] (m\i) to (m\j);
    \else
        \draw[->, bend left=10, blue!50!black, thin] (m\i) to (m\j);
    \fi
    }
  \node[above=2mm,font=\bfseries\small] at (2,4) {Morton (Z-order)};

\end{scope}

\begin{scope}[shift={(7,0)}]
  \foreach \x in {0,...,3} {
    \foreach \y in {0,...,3} {
      \pgfmathtruncatemacro{\blockid}{floor(\x/2) + 2*floor(\y/2)}
      \ifcase\blockid
        \def\thiscolor{red!20}
      \or
        \def\thiscolor{green!20}
      \or
        \def\thiscolor{blue!20}
      \or
        \def\thiscolor{yellow!20}
      \fi

      \filldraw[fill=\thiscolor, draw=black] (\x,\y) rectangle ++(1,1);
      \coordinate (s-\x-\y) at (\x+0.5,\y+0.5);

      \ifodd\y
        \pgfmathtruncatemacro{\snake}{\y*4 + (4-\x)}
      \else
        \pgfmathtruncatemacro{\snake}{\y*4 + \x + 1}
      \fi

      \node (s\snake) at (s-\x-\y) {\snake};
    }
  }

  \foreach \i [evaluate=\i as \j using int(\i+1)] in {1,...,15}
   \draw[->, bend right=20, red!90!black, thin] (s\i) to (s\j);
  \node[above=2mm,font=\bfseries\small] at (2,4) {Snake Order};
\end{scope}
\end{tikzpicture}
\caption{Example of flattening a $4\times4$ grid: Morton (Z-order) and Snake ordering.}
\label{fig:morton-snake}
\end{figure}
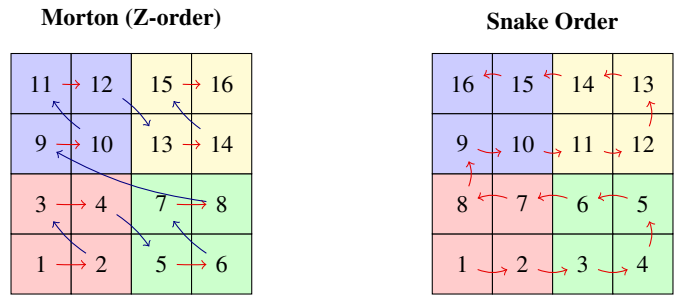

\begin{itemize}
    \item \texttt{MNIST}: 10-way (0-9 digits) classification of a $28 \times 28$ grayscale image of a handwritten digit. The input image is flattened into a $784$-length scalar sequence.
    \item \texttt{CIFAR}~\cite{krizhevsky2009learning}: 10-way image classification using the CIFAR-10 dataset The input is flattened into a sequence of inputs of $1024$-length and three-channel triple (R,G,B). There are $45,000$ training examples, $5,000$ validation examples, and $10,000$ test examples. 
\end{itemize}
Both datasets are flattened in common sequential versions: sMNIST, sCIFAR with snake order, and in quad-tree aligned versions: zMNIST, zCIFAR wth morton order, see Figure \ref{fig:morton-snake}

\texttt{Long Range Arena} (LRA)~\citep{tay2020lra} datasets:
\begin{itemize}
    \item \texttt{Pathfinder}~\cite{linsley2018learning}: a binary classification task.  A $32 \times 32$ grayscale image image shows a start and an end point as a small circle.  There are a number of dashed lines in the image.  The task is to classify whether there is a dashed line (or path) joining the start and end point.  Sequences are all of the same length ($1,024$).  There are $160'000$ training examples, $20'000$ validation examples, and $20'000$ test examples. 

    \item \texttt{Path-X}: Identical to the \texttt{Pathfinder} challenge, with the images are $128 \times 128$ pixels.

    \item \texttt{ListOps}~\cite{nangia2018listops}: 10-way(0-90) digits classification task, representing the integer result of the expression. 
    Given a nested set of mathematical operations (such as \texttt{min} and \texttt{max}) and integer operands in 
    $\{0,\dots 9\}$, compute the integer result of the mathematical expression. Characters are encoded as one-hot vectors, with $17$ unique values possible (opening brackets and operators are grouped into a single token). The sequences are padded to a maximum length of $2,000$ with a fixed indicator value.  There are $96'000$ training sequences, $2,000$ validation sequences, and $2000$ test sequences.  

\end{itemize}

\subsection{Full table}\label{sec:app_fullt}
\begin{table}[!ht]
  \captionsetup{justification=centering}
  \caption{Test accuracy on image classification. We used the table from \cite{smith2023simplified} and add our results.}
  \label{tab:pixellong}
  \centering
  \begin{adjustbox}{center}
    \begin{tabular}{@{}lcccc@{}}
    \toprule
    Model               & \texttt{sMNIST}  & \texttt{sCIFAR} & \texttt{zMNIST} & \texttt{zCIFAR} \\ 
    (Input length)      & (784)            & (1024)          & (784)           & (1024)          \\ \midrule    
    Transformer \citep{trinh2018learning, vaswani2017attention} & 98.9  & 62.2   & - & - \\  \midrule 
    CCNN \citep{romero2022towards}         & \textbf{99.72}  & \textbf{93.08}  & - & - \\
    FlexTCN \citep{romero2021flexconv}     & 99.62           & 80.82           & - & - \\
    CKConv \citep{romero2021ckconv}        & 99.32           & 63.74           & - & - \\
    TrellisNet \citep{bai2018trellis}      & 99.20           & 73.42           & - & - \\
    TCN \citep{bai2018empirical}           & 99.0            & -               & - & - \\  \midrule
    LSTM  \citep{gu2020improving, hochreiter1997long} & 98.9 & 63.01  & - & - \\
    r-LSTM   \citep{trinh2018learning}     & 98.4            & 72.2            & - & - \\
    Dilated GRU \citep{chang2017dilated}   & 99.0            & -               & - & - \\
    Dilated RNN  \citep{chang2017dilated}  & 98.0            & -               & - & - \\
    IndRNN  \citep{li2018independently}    & 99.0            & -               & - & - \\
    expRNN  \citep{lezcano2019cheap}       & 98.7            & -               & - & - \\     
    UR-LSTM   \citep{gu2020improving}      & 99.28           & 71.00           & - & - \\
    UR-GRU \citep{gu2020improving}         & 99.27           & 74.4            & - & - \\
    LMU     \citep{Voelker2019LegendreMU}  & -               & -               & - & - \\
    HiPPO-RNN  \citep{gu2020hippo}         & 98.9            & 61.1            & - & - \\
    UNIcoRNN \citep{rusch2021unicornn}     & -               & -               & - & - \\ 
    LMU-FFT  \citep{chilkuri2021parallelizing} & -            & -               & - & - \\
    LipschitzRNN  \citep{erichson2021lipschitz} & 99.4       & 64.2            & - & - \\  \midrule
    LSSL   \citep{gu2021lssl}              & 99.53           & 84.65           & - & - \\
    S4    \citep{gu2022parameterization, gu2021s4} & 99.63     & 91.80           & - & - \\
    S4D   \citep{gu2022parameterization}   & -               & 89.92           & - & - \\
    Liquid-S4 \citep{hasani2022liquid}     & -               & 92.02           & - & - \\
    S5                                       & 99.65         & 90.10           & 99.5 (our run) & 89.9 (our run) \\ \bottomrule
    \textbf{Myo} (this work) & 99.2 & 92.6 & \textbf{99.7} & \textbf{93.4} \\ \bottomrule
    \end{tabular}
\end{adjustbox}
\end{table}

\end{document}